%% file: main.tex
\documentclass[10pt,twocolumn,letterpaper]{article}

\usepackage{cvpr}              


\definecolor{cvprblue}{rgb}{0.21,0.49,0.74}
\usepackage[pagebackref,breaklinks,colorlinks,allcolors=cvprblue]{hyperref}
\usepackage{tikz}
\usepackage{lipsum}
\usepackage{multirow}
\tikzstyle{mybox} = [draw=black, rectangle]

\usepackage{CJKutf8}
\usepackage{pifont}
\usepackage{amssymb}
\usepackage{latexsym}
\usepackage{booktabs}

\def\paperID{18647} 
\def\confName{CVPR}
\def\confYear{2025}

\title{Relation-Rich Visual Document Generator for Visual Information Extraction}

\author{Zi-Han Jiang\textsuperscript{1}, Chien-Wei Lin\textsuperscript{1}, Wei-Hua Li\textsuperscript{1}, Hsuan-Tung Liu\textsuperscript{2}, Yi-Ren Yeh\textsuperscript{3}, Chu-Song Chen\textsuperscript{1} \\
\textsuperscript{1}National Taiwan University, \textsuperscript{2}E.SUN Financial Holding Co., Ltd., \textsuperscript{3}National Kaohsiung Normal University\\
{\tt\small \{r12922082,r11922212,d12922009,chusong\}@csie.ntu.edu.tw,}\\ {\tt\small ahare-18342@esunbank.com.tw, yryeh@nknu.edu.tw}\\
}

\begin{document}
\maketitle
\input{sec/0_abstract}    
\input{sec/1_intro}

\input{sec/2_related_work}
\input{sec/3_method}
\input{sec/4_experiments}
\input{sec/limitation}
\input{sec/5_conclusion}

\newpage



{
    \small
    \bibliographystyle{ieeenat_fullname}
    \bibliography{main}
}

\input{macros}
\input{sec/X_suppl}

\end{document}

%% file: sec/0_abstract.tex
\begin{abstract}
Despite advances in Large Language Models (LLMs) and Multimodal LLMs (MLLMs) for visual document understanding (VDU), visual information extraction (VIE) from relation-rich documents remains challenging due to the layout diversity and limited training data. While existing synthetic document generators attempt to address data scarcity, they either rely on manually designed layouts and templates, or adopt rule-based approaches that limit layout diversity. 
Besides, current layout generation methods focus solely on 
topological patterns without considering textual content, making them impractical for generating documents with complex 
associations between the contents and layouts. In this paper, we propose a 
\textbf{R}elation-r\textbf{I}ch 
visual \textbf{D}ocument \textbf{GE}nerator (RIDGE) that addresses these limitations through a two-stage approach: (1) Content Generation, which leverages LLMs to generate document content using a carefully designed Hierarchical Structure Text format 
which captures entity categories and relationships, and (2) Content-driven Layout Generation, which learns to create diverse, plausible document layouts solely from easily available Optical Character Recognition (OCR) results, 
requiring no human labeling or annotations efforts. 
Experimental results have demonstrated that our method significantly enhances the performance of document understanding models on various VIE benchmarks.
\end{abstract}

%% file: sec/1_intro.tex
\section{Introduction}
\label{sec:intro}

VDU has attracted significant attention 
in knowledge-centric applications. 
While conventional document pre-trained models ~\cite{xu2020layoutlm,xu-etal-2021-layoutlmv2,huang2022layoutlmv3,peng2022ernie,tu2023layoutmask,hong2022bros} have demonstrated outstanding performance on specific downstream tasks, their practical applicability is often constrained by the necessity of task- or dataset-specific fine-tuning. This has prompted a shift in research focus towards the utilization of 
LLMs and MLLMs.

To perform document analysis, segmented text entities 
are the basic elements in a visual document. Entity Linking (links between entities) and Entity Category (attributes or semantic classification labels of entities) are the two main types of information that are critical to VIE. To learn a document-analysis model, a major difficulty is the lack of labeled training data, where the labels can be of unary relation (containing $n$ entities) or binary relation (containing $O(n^2)$ links between entity pairs). However, the effort for labeling $O(n^2)$ labels is huge, which hinders the development of learning effective models for a comprehensive understanding of the diverse forms of visual documents. 

LLMs and MLLMs possess generalization capability and emergent properties resulting from training on extensive corpora, allowing them to perform effectively on document-related tasks. 
Recent models have shown promising results in document-oriented visual question answering, such as Qwen2-VL~\cite{Qwen2VL} approaching human-level performance on DocVQA~\cite{mathew2021docvqa}. However, significant challenges persist in extracting information from relation-rich (\eg, forms of various formats) or semi-structured documents (\eg, invoices and receipts).
The variability complicates the process of accurately identifying and linking key-value pairs.
A general extraction rule cannot be easily established, affecting model accuracy and requiring extensive training data to enhance the generalization capabilities.

In this paper, we introduce a method to synthesize relation-rich visual documents.
The generated documents exhibit highly diverse layouts, 
arise not only from the document's content but also 
from the 
variability across different document sources. 
Our content-rich generative approach leverages LLMs in two steps. First, we use ChatGPT~\cite{ray2023chatgpt} to generate text with entity-category and entity-link information in a hierarchy. Second, we adopt self-supervised learning to build a model that can place the generated text into appropriate bounding box locations with reasonable layouts. 
Unlike traditional layout generation solutions 
focusing mainly on producing 
bounding boxes without 
relationally meaningful 
textual content~\cite{tang2024layoutnuwa, layoutgan, jiang2023layoutformer++, inoue2023layoutdm}, our generated content-and-relation-rich 
documents can boost the performance of VDU. We demonstrate an overview of existing datasets and our approach as follows.

\noindent \textbf{VIE Datasets.} Existing datasets for 
VIE falls into two categories: open-set and closed-set. Open-set datasets provide annotations for entity-level categories and entity linking. For example, FUNSD~\cite{jaume2019funsd}, an English form-type document dataset, classifies entities into four categories (header, question, answer, and other), and also 
provides links between questions and answers. In contrast, closed-set datasets define a set of specific keys and extract corresponding values. For example, CORD~\cite{park2019cord}, a consolidated receipt dataset, includes 30 categories, such as menu name and total price. 
Compared with closed-set datasets, the number of images in open-set datasets is quite limited, as shown in \cref{table:dataset-statistics}. For instance, FUNSD contains only 199 document images, while XFUND~\cite{xu-etal-2022-xfund}, a form-type document dataset covering 7 languages, also has only 199 document images per language. Additionally, data collection poses another challenge, as these types of documents may contain personal information, making them difficult to access publicly. This shortage further highlights the pressing need for a tool that can automatically generate large volumes of this type of data, complete with annotations and document images.

\input{sec/table_dataset}

\noindent\textbf{Overview of Our Approach.} 
We propose RIDGE, a relation-rich visual document generator 
capable of producing 
meaningful document content that inherits complex hierarchical relationships between entities, while maintaining a highly diverse layout. To achieve this, we introduce a two-stage generation pipeline. \textbf{\uppercase\expandafter{\romannumeral 1}. Content Generation.} 
In this work, we primarily focus on synthesizing open-set VIE data to address its scarcity. 
We leverage LLMs to generate document content in our meticulously designed \textit{Hierarchical Structure Text (HST)} format, where the generated output can be directly parsed into segmented text entities with their category and linking annotations. 
With content and annotations now prepared, we advance to the next stage. \textbf{\uppercase\expandafter{\romannumeral 2}. Content-driven Layout Generation.} Research~\cite{lin2023layoutprompter,tang2024layoutnuwa,feng2023layoutgpt,cho2024visual} has shown that LLMs demonstrate layout planning abilities
attributed to their training on layout-specific sources like HTML code of mobile app interfaces or websites. Inspired by this, we leverage LLMs to generate document layouts 
through self-supervised learning. 
Our training data requires only OCR results (segmented text entities and their bounding box coordinates) from documents that are readily obtainable through off-the-shelf OCR tools.
Although we do not provide entity category or linking annotations to the model, it 
can independently learn content-layout relationships through 
self-supervision, generating a reasonable layout that aligns with the annotations produced in the previous stage. 
Additionally, we introduce a training paradigm called \textit{Hierarchical Structure Learning} that further leverages HST for reinforcing document understanding models' comprehension.
Experimental results have shown that RIDGE significantly enhances the performance of document understanding models on various VIE benchmarks.


    

%% file: sec/table_dataset.tex
\begin{table}[!t]
\centering
\tabcolsep=0.07cm
\resizebox{1\columnwidth}{!}{
    
    \begin{tabular}{ccc|ccccc}
\hline
         & \multicolumn{2}{c|}{\textbf{Open-set}}                                    & \multicolumn{5}{c}{\textbf{Closed-set}}                                                                \\ \hline
Dataset  & \begin{tabular}[c]{@{}c@{}}FUNSD\\ \cite{jaume2019funsd}\end{tabular} & \begin{tabular}[c]{@{}c@{}}XFUND\\ \cite{xu-etal-2022-xfund}\end{tabular}     & \begin{tabular}[c]{@{}c@{}}CORD\\ \cite{park2019cord}\end{tabular} & \begin{tabular}[c]{@{}c@{}}EPHOIE\\ \cite{wang2021towards}\end{tabular} & \begin{tabular}[c]{@{}c@{}}POIE\\ \cite{kuang2023visual}\end{tabular} & \begin{tabular}[c]{@{}c@{}}SROIE\\ \cite{huang2019icdar2019}\end{tabular} & \begin{tabular}[c]{@{}c@{}}DocILE\\ \cite{docile}\end{tabular}          \\ \hline
\#Img. & 199   & \begin{tabular}[c]{@{}c@{}}199 \\ per lang.\end{tabular} & 1000 & 1494   & 3000 & 1000  & \begin{tabular}[c]{@{}c@{}}real: 6680\\ syn:100K*\end{tabular} \\ \hline
\#Keys   & Free  & Free                                                     & 30   & 10     & 21   & 4     & 55                                                             \\ \hline
\end{tabular}
}
\caption{Dataset statistics of VIE tasks. * represents the synthetic data are generated by 100 templates.}
\vspace{-15pt}
\label{table:dataset-statistics}
\end{table}

%% file: sec/2_related_work.tex
\definecolor{color_header}{RGB}{237,196,241}
\definecolor{color_key}{RGB}{224,187,190}
\definecolor{color_value}{RGB}{202,219,235}
\setlength{\fboxsep}{2pt} 
\begin{figure*}[!t]
\centering
\includegraphics[width=1.\linewidth]{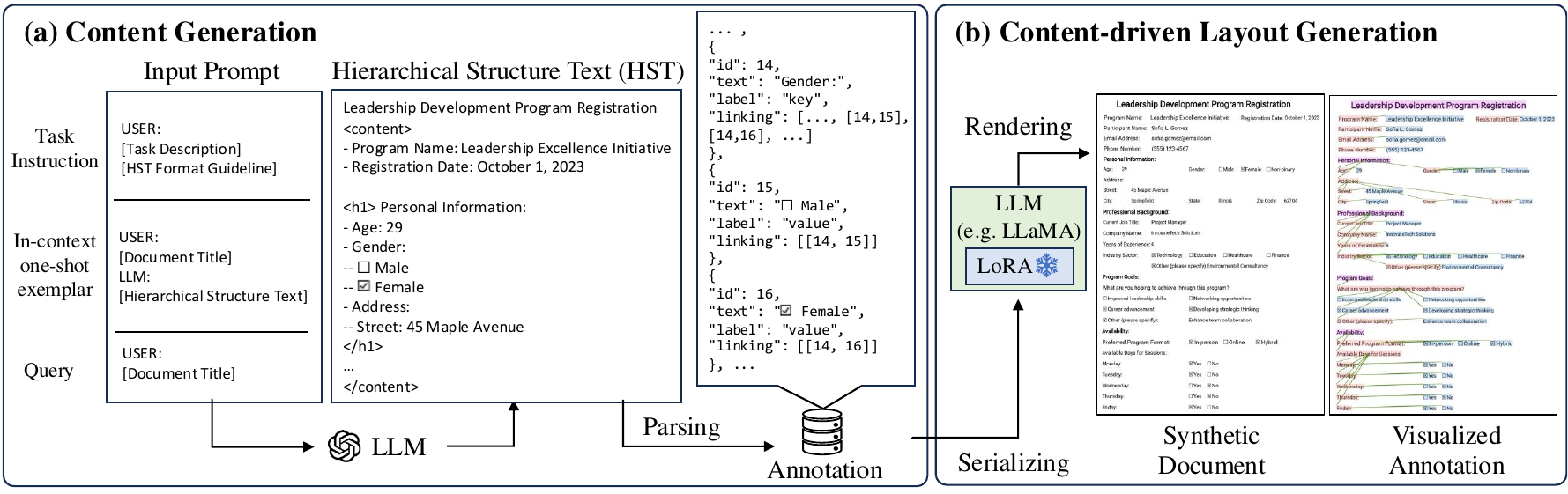}
  \caption{Overview of RIDGE, including (a) Content Generation and (b) Content-driven Layout Generation. In the visualized annotation, purple represents the \colorbox{color_header}{header}, red represents the \colorbox{color_key}{key}, blue represents the \colorbox{color_value}{value}, and green lines represent entity linking.}
  \vspace{-15pt}
  \label{fig:overview}
\end{figure*}

\section{Related Work}
\label{sec:related_work}

\textbf{Synthetic Document Generators.} Due to the high cost of manual annotation and the limited availability of sensitive documents, recent research has focused on automatically synthesizing documents to meet the substantial demands of machine learning applications. DocSynth~\cite{biswas2021docsynth} uses a Generative Adversarial Network (GAN) to generate synthetic document images based on manually-designed layouts. 
Due to the low resolution ($128\times128$) of the generated images, the method can generate only blurry shadows for text and
fails to capture detailed document content, 
resulting 
the generated synthetic documents 
only suitable for document layout analysis. SynthDoG~\cite{kim2022donut} aims to generate documents for OCR training by selecting text content from Wikipedia at random and rendering it onto a natural or document-textured image. Its layouts are generated with a rule-based algorithm that merely 
places random grids on the image, causing content to split across grids in ways that may appear unnatural. DocILE~\cite{docile} generates realistic content with 
key information annotations; however, its synthetic method is rule-based, relying on only 100 annotated layout templates. It primarily alters content or applies style changes to font and borders, which limits its diversity and suitability for general 
training. Existing methods are limited in their ability to produce realistic content and diverse layouts, both of which are essential for robust VIE training.

\noindent\textbf{Document Layout Generation.} Automatic layout generation has emerged as a crucial task across diverse application domains, spanning documents, natural images, user interfaces, and posters. Early studies~\cite{layoutgan,read, jyothi2019layoutvae, arroyo2021VTN, jiang2022coarse} employ GANs and VAEs to learn layout generation through reconstruction, while Jiang \etal~\cite{jiang2023layoutformer++} reformulate this as a sequence-to-sequence problem using transformer architectures. Through advances in diffusion techniques, some studies~\cite{inoue2023layoutdm, hui2023unifying, he2023diffusionbased} demonstrate the effectiveness of treating layout generation as a diffusion process. More recently, researchers~\cite{tang2024layoutnuwa, lin2023layoutprompter, feng2023layoutgpt, cho2024visual, lian2024llmgrounded} have begun to utilize LLMs for this task, harnessing their inherent layout understanding. 
While most methods focus on generating layout (bounding box and category) independent of content, some~\cite{fengheng2023relation, jia2024colehierarchicalgenerationframework, textdiffuser2} have incorporated text content with layout planning, primarily focusing on visual design contexts such as posters and graphic designs. These approaches typically handle sparse text elements (keywords or short phrases) arranged around visual components, rather than documents with dense textual content and complex structural relationships. Among document-layout-related research, most studies predominantly concentrate on academic paper formats due to the accessibility of training data. These methods address solely layout generation, without considering the specific textual content to populate these layouts.
Given that font sizes in documents typically follow certain alignment principles, applying such pre-generated layouts would restrict users to fixed-length text within each layout bounding box—a constraint that makes these methods impractical for generating documents with meaningful textual content.

Consequently, existing studies can only be applied to training document layout analysis models, rather than directly supporting VIE tasks, as the latter requires models to develop robust comprehension of complex content-layout relationships. 
Moreover, previous methods rely heavily on supervised training with layout element categories, which is impractical for relation-rich documents due to limited publicly available annotations. In contrast, our proposed self-supervised method requires only readily available OCR results for training, making it more practical and scalable for real-world applications.

%% file: sec/3_method.tex
\begin{figure*}[!ht]
\centering
\includegraphics[width=1\linewidth]{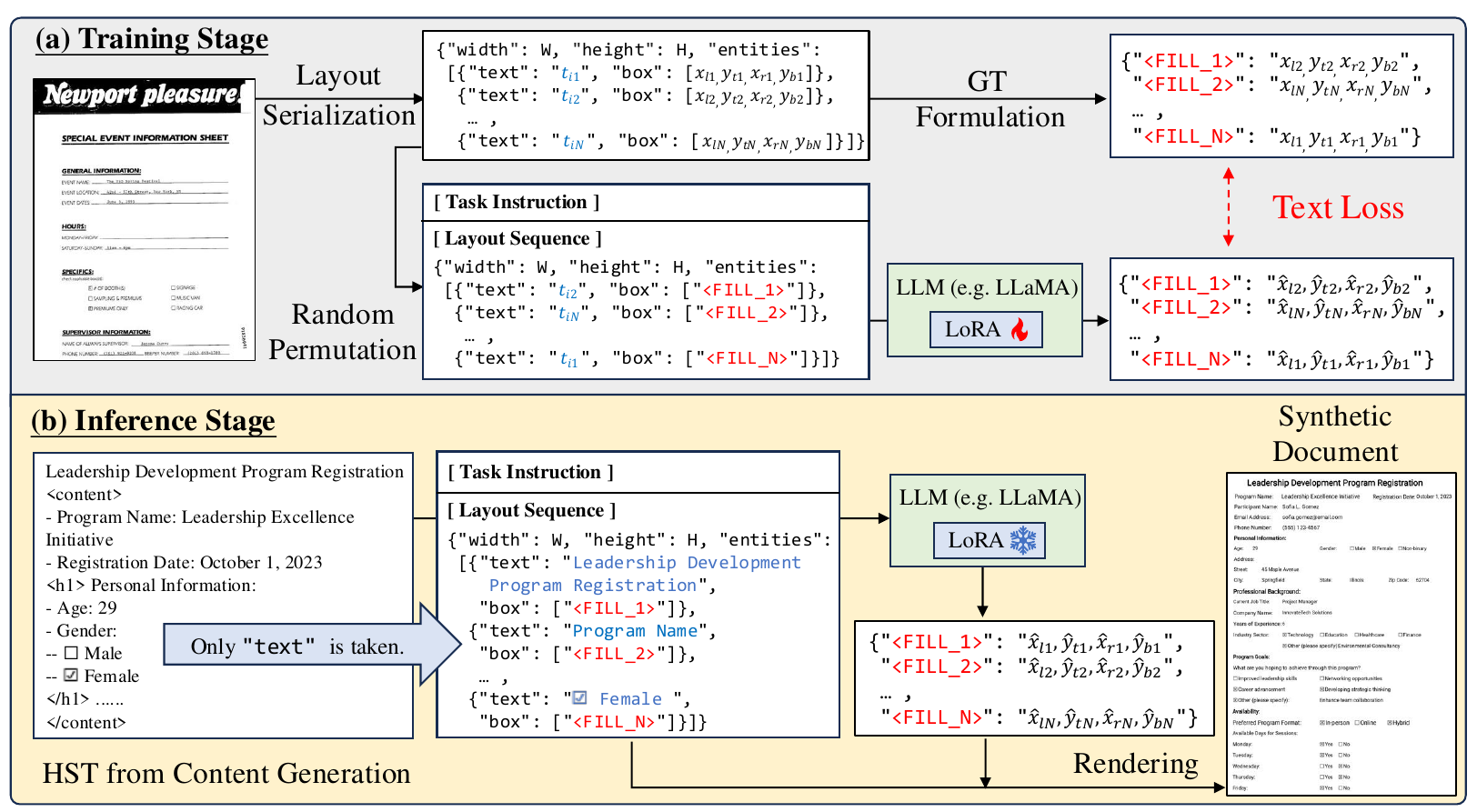}
  \caption{Content-driven Layout Generation Model (CLGM).}
  \vspace{-10pt}
  \label{fig:clgm}
\end{figure*}

\section{Method}
\label{sec:method}

This section elaborates on RIDGE, a novel method for 
synthesizing documents with realistic content and diverse layouts. It employs a two-stage pipeline: \textbf{content generation} for text generation and annotation parsing, and \textbf{content-driven layout generation} for spatial arrangement of document entities, as shown in \cref{fig:overview}.

In this work, we aim to generate open-set VIE dataset consisting of document images and their corresponding annotations, including entity categories and linking, similar to FUNSD. Formally, a document $D = \{e_i\}_{i=1}^N$ is composed of entities, where $N$ is the number of entities in the document. Each entity $e = (id, t, b, c, l)$, where $id$ is the entity’s unique identifier, $t$ denotes its textual content, $b = (x_\text{left}, y_\text{top}, x_\text{right}, y_\text{bottom})$ specifies its bounding box coordinates, $c \in \{\text{header}, \text{key}, \text{value}, \text{other}\}$ indicates its semantic category; and $l$ contains a list of links with other entities, where a link is represented as $[id_\text{key}, id_\text{value}]$.

\subsection{Content Generation}
\label{subsec:content_generation}

As shown in \cref{fig:overview}(a), given a document title, LLMs generate corresponding detailed content of the document in the format of Hierarchical Structure Text (HST). The format is structured as follows: (1) the entire document is wrapped within \texttt{<content>} tags after the document title, (2) paragraphs are organized using header tags (\texttt{<h1>}, \texttt{<h2>}, etc.), where text immediately following each header tag is treated as the paragraph’s title, and (3) the content is organized in a tree structure through two patterns—using colon (':') for direct key-value pairs in the same line (e.g., "Name: John"), and using different numbers of hyphens ('-') to indicate hierarchical nesting relationships.

This structured format enables automatic extraction of entity text, categories, and linking relationships. To help LLMs get a more comprehensive understanding of HST, we give a detailed task instruction and an in-context one-shot exemplar for LLMs to reference. For detailed prompts, please refer to the supplementary material.

\subsection{Content-driven Layout Generation (CLGM)} 
In this subsection, we introduce CLGM (see \cref{fig:clgm}), which is capable of arranging text segments to proper bounding box positions. Instead of traditional category-driven layout generation methods~\cite{tang2024layoutnuwa, inoue2023layoutdm}, our model is content-driven to tackle nuanced text content in diverse documents.
CLGM consists of three parts: \textit{serializing the document layout}, \textit{self-supervised layout learning}, and \textit{document rendering}.

\subsubsection{Document Layout Serialization}

Since LLMs accept only text input, we need to represent a document layout as a sequence. Previous studies~\cite{tang2024layoutnuwa, lin2023layoutprompter, feng2023layoutgpt} have demonstrated that LLMs exhibit strong comprehension in programming languages, due to the inclusion of code-based data in their training corpora. Thus, the layout of a document $D$ can be represented in JSON format as:


\vspace{2pt}\begin{tikzpicture}
\hspace{-10pt}
\node [mybox] (box){
\begin{minipage}{.98\linewidth}
\texttt{\{"width":W,"height":H,"entities":[}
\newline
\texttt{\{"text":}$t_1$\texttt{,"box":}[$x_{\text{left}_1}$, $y_{\text{top}_1}$, $x_{\text{right}_1}$, $y_{\text{bottom}_1}$]\texttt{\}}, ..., 
\newline
\texttt{\{"text":}$t_N$\texttt{,"box":}[$x_{\text{left}_N}$,$y_{\text{top}_N}$,$x_{\text{right}_N}$,$y_{\text{bottom}_N}$]\texttt{\}]\}}
\end{minipage}};
\end{tikzpicture}


\noindent where $W$ and $H$ indicate the width and height of the layout canvas, and \textit{"entities"} consists of all entities $e_i$ that comprise document $D$. Unlike previous methods~\cite{tang2024layoutnuwa, lin2023layoutprompter, feng2023layoutgpt} that use HTML or CSS for layout representation, we adopt JSON to reduce input sequence length by eliminating redundant attribute text. In HTML and CSS, an element's bounding box requires repeating attribute text. 
Our representation consolidates attributes into a single ``box" field, which particularly reduces sequence length for complex document layouts containing hundreds of entities.

\subsubsection{Layout Self-Supervised Learning}

To generate layouts for documents with varying content, we formulate this task as predicting bounding box coordinates for each given text entity. 
Our approach is inspired by LayoutNUWA~\cite{tang2024layoutnuwa}, a method purely for layout generation without text contents. Our method is content-driven and performs self-supervised learning in a more general way for placing the texts into suitable positions. 
We use a masking strategy where bounding box coordinates are replaced with special mask tokens \texttt{<FILL\_i>} for each $e_i$. Specifically, let $S(D_{\backslash M})$ denote the input masked sequence to model $f_{\theta}$:

\vspace{2pt}\begin{tikzpicture}
\hspace{5pt}
\node [mybox] (box){
\centering
\begin{minipage}{.85\linewidth}
\texttt{\{"width":W,"height":H,"entities":[}
\newline
\texttt{\{"text":}$t_1$\texttt{,"box":[}\texttt{"\textless FILL\_1\textgreater"]\},} ..., 
\newline
\texttt{\{"text":$t_N$,"box":["\textless FILL\_N\textgreater"]\}]\}}
\end{minipage}};
\end{tikzpicture}

\noindent where $S(\cdot)$ is a serialization function and $M$ denotes the masked bounding boxes. Model $f_{\theta}$ then output a JSON dictionary sequence $S(M)$ that directly maps each mask token to its corresponding coordinates:

\vspace{2pt}\begin{tikzpicture}
\hspace{4pt}
\node [mybox] (box){
\centering
\begin{minipage}{.9\linewidth}
\texttt{\{"\textless FILL\_1\textgreater":} ``$x_{\text{left}_1}$, $y_{\text{top}_1}$, $x_{\text{right}_1}$, $y_{\text{bottom}_1}$"\texttt{,} ..., 
\newline
\makebox[2pt][l]{}
\texttt{"\textless FILL\_N\textgreater":} "$x_{\text{left}_N}$, $y_{\text{top}_N}$, $x_{\text{right}_N}$, $y_{\text{bottom}_N}$"\texttt{\}}
\end{minipage}};
\end{tikzpicture}
Content-based layout generation can thus be written as
\begin{equation}
  S(M) = f_{\theta}(S(D_{\backslash M})).
  \label{eq:CLGM_formulation}
\end{equation}
Unlike LayoutNUWA which requires LLMs to complete a code template by repeating the entire input with predicted coordinates, our method streamlines this process by focusing solely on predicting bounding box coordinates. This 
reduces the output sequence length 
and enhances the efficiency 
for layout generation.
Following the standard auto-regressive formulation, we optimize our model using the cross-entropy loss:
\begin{equation}
  \mathcal{L} = - \sum_{k=1}^{K} \log P(S(M)^k | S(M)^{<k}, S(D_{\backslash M}), \theta),
  \label{eq:loss}
\end{equation}
with $K$ the number of tokens in the output sequence $S(M)$, $S(M)^k$ the $k^{th}$ token in $S(M)$, and $\theta$ the model parameters.

Considering the auto-regressive nature of LLMs, we randomly permute the order of input entities during training to prevent the model from learning a fixed layout order, as the document layout inherently has no natural sequence. 
This randomization strategy brings a crucial benefit—by exposing LLMs to highly unordered text content, which is even challenging for humans to reconstruct, 
the model learns to infer reasonable relationships between entities solely from their text. This allows our method to generate plausible layouts without requiring category and linking annotations, thus eliminating the need for human annotation effort. 
During inference, we maintain the order of entities as they appear in HST (\ie, human reading order) to facilitate the model's understanding of document content. 

Additionally, we train our model with non-normalized canvas sizes, enabling it to handle diverse layout dimensions directly. 
It enhances the model's adaptability in 
practice where documents can vary significantly in size.

\subsubsection{Document Rendering}

After obtaining the layout coordinates from CLGM, the document rendering module transforms the layout sequence into a visually coherent document image. First, we 
validate the LLM output to ensure proper text rendering within bounding boxes. Second, we enhance the visual appearance 
by various styling operations to create realistic documents.

\noindent\textbf{LLM output processing and Styling.} 
When using LLMs to generate structured output, format validation is inevitable. 
We verify that CLGM's output conforms to valid JSON format and contains all expected mask tokens, 
excluding samples that fail these criteria. 
Subsequently, we render text from the left boundary of each bounding box, with font size determined by the box height generated by CLGM. Since different fonts may yield varying text widths, we adaptively extend the box width and shift subsequent entities in the same row when text exceeds its allocated bounds, preserving readability and CLGM's overall layout design.
In addition, we enhance the visual appearance through 
changing fonts, applying bold 
formatting to text labeled as header, adding different background materials, and incorporating grid lines around entities. 

\input{sec/table_imp_zero_vie}


\begin{figure}[!t]
\centering
\includegraphics[width=1.\linewidth]{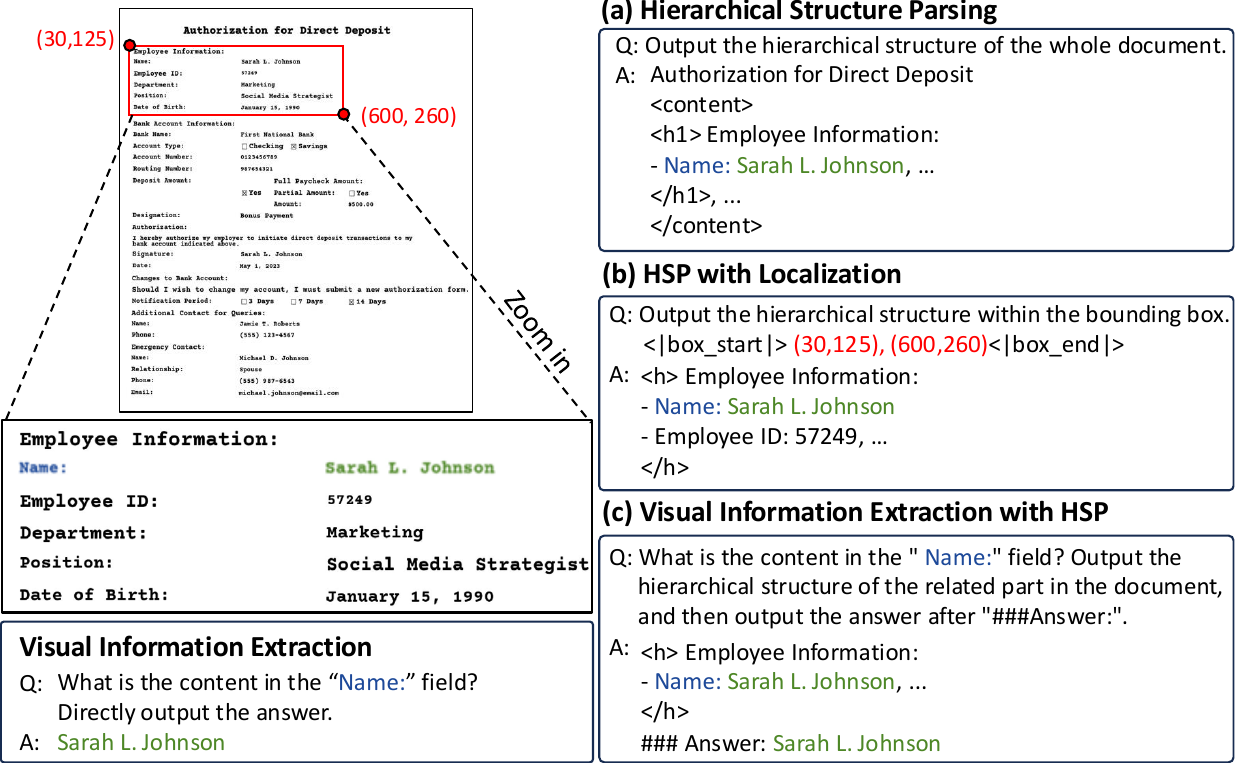}
  \caption{Hierarchical Structure Learning}
  \vspace{-15pt}
  \label{fig:hsl}
\end{figure}

\subsection{Hierarchical Structure Learning}
Our generated documents with annotations 
can be directly applied to any document understanding method for training. 
W.L.O.G., we use MLLMs~\cite{Qwen2VL, liu2024llavanext} for training due to their promising performance recently. To exploit our generated HST better, we propose a training paradigm that extensively utilizes HST to enhance models' comprehension. Existing LLM/MLLM-based methods employ text-layout reconstruction learning to enhance models' understanding of document structure. For example, DocOwl-1.5~\cite{docowl1.5} requires models to output the document text with newline characters ('\textbackslash n') and white spaces inserted to mimic the layout of the document image. Additionally, LayoutLLM~\cite{layoutllm} requires models to reconstruct the document by generating a sequence of entities in the format of ``\textless\{box\}, \{text\}\textgreater". While these tasks help models learn to associate text with its spatial positions, they may not capture the complex hierarchical relationships inherent in document layouts. To address this issue, we 
introduce Hierarchical Structure Learning using HST generated by RIDGE to strengthen models’ comprehension of hierarchical relationships within document text and layout. 

\noindent\textbf{Hierarchical Structure Parsing (HSP)} requires models to parse the entire document content into our HST format. As shown in \cref{fig:hsl}(a), given a document image, models need to identify paragraph boundaries, understand the hierarchical nesting relationships within its content, and output HST with proper tags and indentation levels. This task helps models understand both the overall structure and complex relationships of the document.

\noindent\textbf{HSP with Localization} is built upon the basic HSP task. We introduce spatial awareness by requiring models to parse hierarchical structures within specific regions. As shown in \cref{fig:hsl}(b), models are given a region represented by bounding box coordinates (normalized to 0-999), and are required to output HST for content within that region. This task maintains the localization ability of pre-trained models during downstream training while facilitating hierarchical structure learning by reducing the task to a smaller region.

\noindent\textbf{Visual Information Extraction with HSP} combines HSP with VIE task. As shown in \cref{fig:hsl}(c), when querying specific information (e.g., "Name:"), models first output the hierarchical structure of the relevant document section, then provide the answer after “\#\#\#Answer:”. Inspired by the idea of Chain-of-Thought~\cite{cot}, this approach encourages models to understand document context through its hierarchical structure before extracting information, potentially leading to more robust and interpretable results.


%% file: sec/table_imp_zero_vie.tex
\begin{table*}[!ht]
\centering
\renewcommand{\arraystretch}{0.98}
\tabcolsep=0.08cm
    {\resizebox{0.9\linewidth}{!}{
    \begin{tabular}{l|cc|ccccccc}
    \toprule
    & \multicolumn{2}{c|}{Open-set} & \multicolumn{7}{c}{Closed-set}\\ \cline{2-10} 
    & \multicolumn{1}{c|}{\textbf{FUNSD}} & \textbf{XFUND-ZH} & \multicolumn{1}{c|}{\textbf{CORD}} & \multicolumn{1}{c|}{\textbf{CORD\textsuperscript{--}}} & \multicolumn{2}{c|}{\textbf{EPHOIE}} & \multicolumn{2}{c|}{\textbf{POIE}} & \textbf{SROIE\textsuperscript{--}} \\ 
    \multirow{-3}{*}{Methods}             & \multicolumn{1}{c|}{\textit{F1 \%}}                                & \textit{F1 \%}                                & \multicolumn{1}{c|}{\textit{F1 \%}}                                & \multicolumn{1}{c|}{\textit{ANLS \%}}                              & \textit{Acc \%}                               & \multicolumn{1}{c|}{\textit{ANLS \%}}                              & \textit{Acc \%}                               & \multicolumn{1}{c|}{\textit{ANLS \%}}                              & \textit{ANLS\%}                                \\ \hline
    LayoutLLM~\cite{layoutllm} \textit{(CVPR’24)}                      & \multicolumn{1}{c|}{-}                                & -                                & \multicolumn{1}{c|}{-}                                & \multicolumn{1}{c|}{63.10}                                & -                               & \multicolumn{1}{c|}{-}                                & -                                & \multicolumn{1}{c|}{-}                                & 72.12                                 \\
    Monkey~\cite{li2024monkey} \textit{(CVPR’24)}                      & \multicolumn{1}{c|}{34.27}                                & 28.02                                & \multicolumn{1}{c|}{54.54}                                & \multicolumn{1}{c|}{69.48}                                & 22.44                                & \multicolumn{1}{c|}{30.96}                                & 30.64                                & \multicolumn{1}{c|}{43.95}                                & 64.95                                 \\
    DocOwl-1.5-Chat~\cite{docowl1.5} \textit{(EMNLP’24)}           & \multicolumn{1}{c|}{50.88}                                & 10.14                                & \multicolumn{1}{c|}{64.01}                                & \multicolumn{1}{c|}{63.15}                                & 4.85                                 & \multicolumn{1}{c|}{8.25}                                 & 51.57                                & \multicolumn{1}{c|}{61.35}                                & 61.09                                 \\ \hline
    LLaVA-NeXT-Mistral-7B~\cite{liu2024llavanext}       & \multicolumn{1}{c|}{31.18}                                & 6.02                                 & \multicolumn{1}{c|}{46.58}                                & \multicolumn{1}{c|}{63.81}                                & 4.31                                 & \multicolumn{1}{c|}{8.02}                                 & 49.88                                & \multicolumn{1}{c|}{60.05}                                & 58.59                                 \\
    LLaVA-NeXT-Mistral-7B+               & \multicolumn{1}{c|}{33.41}                                & 10.33                                & \multicolumn{1}{c|}{52.32}                                & \multicolumn{1}{c|}{63.93}                                & 5.39                                 & \multicolumn{1}{c|}{12.07}                                & 51.46                                & \multicolumn{1}{c|}{62.13}                                & 58.58                                 \\
    {\color[HTML]{3531FF} \textbf{$\Delta\uparrow$}} & \multicolumn{1}{c|}{{\color[HTML]{3531FF} \textbf{2.23}}} & {\color[HTML]{3531FF} \textbf{4.31}} & \multicolumn{1}{c|}{{\color[HTML]{3531FF} \textbf{5.74}}} & \multicolumn{1}{c|}{{\color[HTML]{3531FF} \textbf{0.12}}} & {\color[HTML]{3531FF} \textbf{1.08}} & \multicolumn{1}{c|}{{\color[HTML]{3531FF} \textbf{4.05}}} & {\color[HTML]{3531FF} \textbf{1.58}} & \multicolumn{1}{c|}{{\color[HTML]{3531FF} \textbf{2.08}}} & {\color[HTML]{3531FF} \textbf{-0.01}} \\ \hline
    Qwen2-VL-7B~\cite{Qwen2VL}            & \multicolumn{1}{c|}{59.89}                                & 62.08                                & \multicolumn{1}{c|}{82.71}                                & \multicolumn{1}{c|}{80.40}                                & 76.91                                & \multicolumn{1}{c|}{86.52}                                & 93.51                                & \multicolumn{1}{c|}{96.01}                                & 97.50                                 \\
    Qwen2-VL-7B+                         & \multicolumn{1}{c|}{66.48}                                & 69.84                                & \multicolumn{1}{c|}{84.47}                                & \multicolumn{1}{c|}{85.53}                                & 77.89                                & \multicolumn{1}{c|}{87.79}                                & 94.04                                & \multicolumn{1}{c|}{96.71}                                & 97.74                                 \\
    {\color[HTML]{3531FF} \textbf{$\Delta\uparrow$}} & \multicolumn{1}{c|}{{\color[HTML]{3531FF} \textbf{6.59}}} & {\color[HTML]{3531FF} \textbf{7.76}} & \multicolumn{1}{c|}{{\color[HTML]{3531FF} \textbf{1.76}}} & \multicolumn{1}{c|}{{\color[HTML]{3531FF} \textbf{5.13}}} & {\color[HTML]{3531FF} \textbf{0.98}} & \multicolumn{1}{c|}{{\color[HTML]{3531FF} \textbf{1.27}}} & {\color[HTML]{3531FF} \textbf{0.53}} & \multicolumn{1}{c|}{{\color[HTML]{3531FF} \textbf{0.70}}} & {\color[HTML]{3531FF} \textbf{0.24}} \\
    \bottomrule
    \end{tabular}}}
\caption{Performance improvements in zero-shot VIE for MLLMs fine-tuned with RIDGE. "+" denotes that the model is extensively trained with RIDGE. $\Delta\uparrow$ represents performance gain.}
\vspace{-10pt}
\label{tab:MLLM+syn}
\end{table*}

%% file: sec/4_experiments.tex
\section{Experiments}
\label{sec:experiments}
\input{sec/table_comb_RIDGE}

We present the experimental setups for RIDGE, including implementation details, evaluation setup, and results.

\subsection{Implementation Details}
\noindent\textbf{Models.}
For the LLM in the content generation stage, we adapt GPT-4o-mini~\cite{ray2023chatgpt} to generate various document-related content with themes spanning business, education, government, and medical use in both English and Chinese. As for CLGM, we use LLaMA-3.1-8B~\cite{dubey2024llama} as our backbone and fine-tune it with LoRA~\cite{hu2022lora}.
The max sequence length is set as 8000 to accommodate the long layout sequence formed by hundreds of layout entities. Content generation prompts and more training setup are in the supplementary. 
\noindent\textbf{Datasets.}
To develop CLGM’s expertise in form-like document layout design, we collect approximately 100K document images with OCR annotations from publicly available VDU datasets. The majority of our training data originates from RVL-CDIP~\cite{harley2015evaluation}, a document classification dataset comprising 16 classes with 25K images per class. We specifically select the ``form", ``specification", ``resume", and ``memo" classes for training due to their highly structured layouts.  Additionally, we incorporate form-type VIE datasets, including FUNSD~\cite{jaume2019funsd}, XFUND~\cite{xu-etal-2022-xfund} with 7 languages. We also utilize HUST-CELL~\cite{hustcell} 
and manually select approximately 600 structurally complex document images from it. For FUNSD, XFUND, and HUST-CELL, only images from the official training sets are used. Regarding OCR annotations, we use the results provided by the datasets themselves; while RVL-CDIP does not provide OCR annotations, we use Google Vision API to obtain OCR results. Given that FUNSD and XFUND consist entirely of form documents, we treat them as high-quality data and apply 4 times up-sampling during training.

\subsection{Evaluation Setup}
Since RIDGE is the first work that can automatically generate relation-rich visual documents with annotations, and the layout evaluation 
criteria used in previous studies~\cite{tang2024layoutnuwa, inoue2023layoutdm, jiang2023layoutformer++} do not consider text content in the document, it is difficult to apply those metrics (such as FID) to evaluate our work. Thus, 
we conduct comprehensive experiments applying RIDGE on MLLMs as well as the widely used document pre-trained model, LayoutLMv3~\cite{huang2022layoutlmv3}. We assess the performance gains brought by RIDGE on various VIE benchmarks, including FUNSD, XFUND-ZH, CORD, CORD\textsuperscript{--}, EPHOIE, POIE, and SROIE\textsuperscript{--}. Benchmarks\textsuperscript{--} are the cleaned version used in LayoutLLM~\cite{layoutllm}, where keys with multiple values are filtered out. The synthetic documents (referred to as RIDGE) used in the 
experiments comprise approximately 
3K English and 3K Chinese documents, yielding 444K instruction samples, including 194K classical VIE and additional Hierarchical Structure Learning samples.
Examples of generated images are shown in ~\cref{fig:example}(a)
, with more in the supplementary material.

\begin{figure}[!t]
\centering
\includegraphics[width=1.\linewidth]{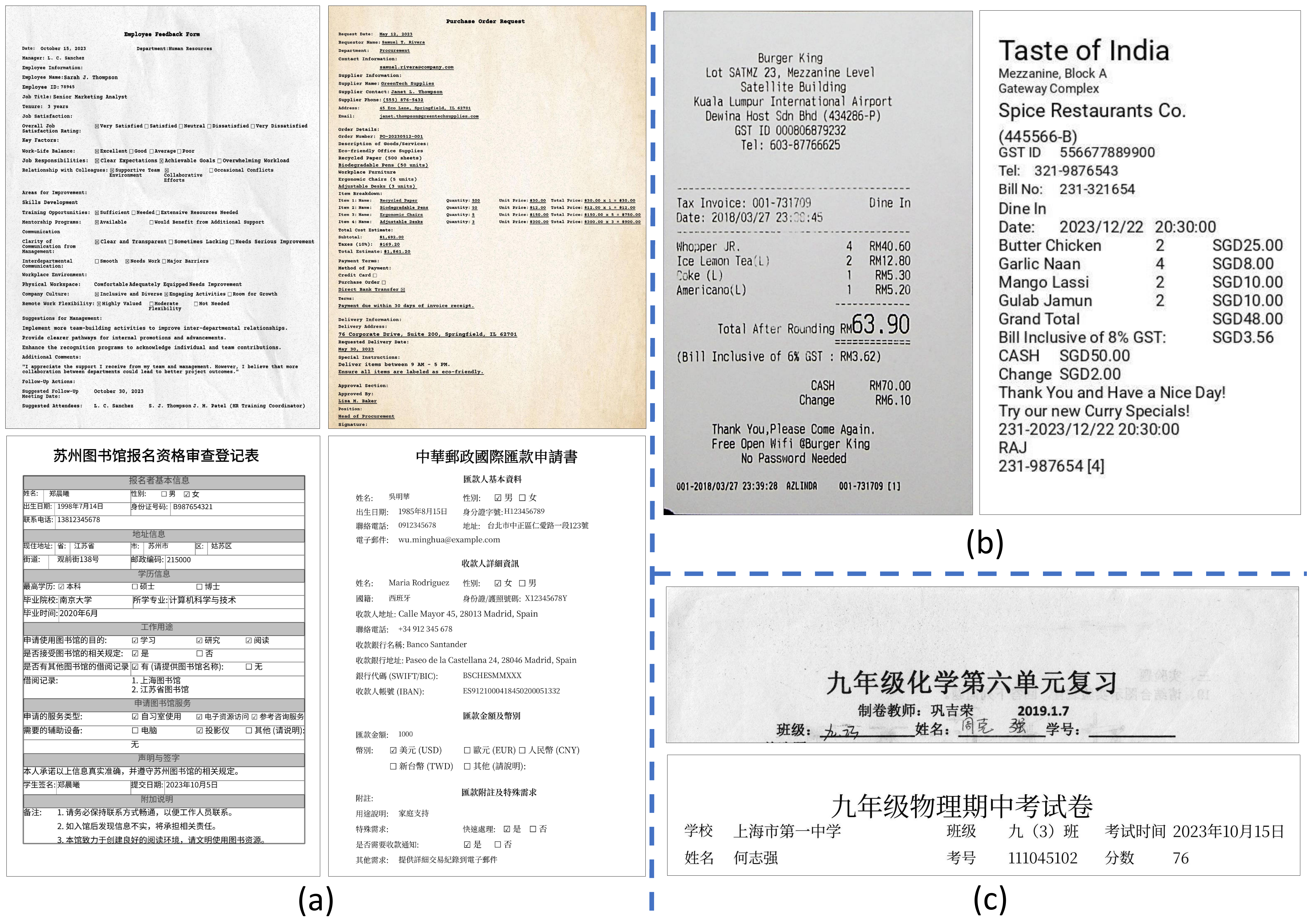}
  \caption{Example of generated documents. (a) General form-like images. (b) Right: SROIE-styled image; Left: real SROIE image. (c) Bottom: EPHOIE-styled image; Top: real EPHOIE image.}
  \vspace{-15pt}
  \label{fig:example}
\end{figure}


\subsection{Fine-tuning MLLMs with RIDGE}
\label{subsec:finetune_MLLMs}
In this subsection, we evaluate the performance gains of MLLMs after being extensively fine-tuned with RIDGE. To strengthen the hierarchical structure understanding of relation-rich documents, we apply 
Hierarchical Structure Learning during training. 
Experiments are conducted on LLaVA-NeXT-Mistral-7B (referred to as LLaVA-NeXT) and Qwen2-VL-7B (referred to as Qwen2-VL). 
More details on training are provided in the supplementary material.

\noindent\textbf{Fine-tuning with RIDGE alone.} As shown in \cref{tab:MLLM+syn}, RIDGE consistently enhances model performance across all evaluation benchmarks, with the only exception being a marginal decrease of 0.01\% for LLaVA-NeXT on SROIE\textsuperscript{--}. While LLaVA-NeXT initially scores behind DocOwl-1.5~\cite{docowl1.5} on some benchmarks, which is an MLLM mainly trained on VDU tasks, extensive training with RIDGE enables it to surpass DocOwl-1.5. Regarding Qwen2-VL, RIDGE successfully pushes its performance to a new state-of-the-art level compared with previous studies.

Further analyzing the results across different benchmark types, we observe that the performance gains are more substantial in open-set VIE benchmarks (exceeding more than 6\%) compared to closed-set benchmarks, particularly for Qwen2-VL. This aligns with our expectations, as RIDGE is mainly designed to generate documents that emphasize open-set scenarios, which promotes better generalization. Beyond its primary strength in open-set data, training on these generalized key-value pairs also yields performance improvements on closed-set data. This is significant because closed-set data often contains inherent biases due to domain-specific key definitions that typically require extensive context to interpret correctly.

\noindent\textbf{Fine-tuning with existing real-world datasets.}
To further validate the efficacy of RIDGE in diverse training scenarios, we conduct experiments using RIDGE independently and in combination with real-world datasets. For a comprehensive evaluation, we incorporate DocVQA, a widely used 
dataset containing human-annotated QA pairs for general VDU training, and FUNSD, an open-set VIE dataset with annotated form-type documents. We use RIDGE-generated English documents in this experiment and evaluate the performance gains on English benchmarks. The results of various dataset combinations—(1) RIDGE alone, (2) RIDGE + DocVQA, and (3) RIDGE + DocVQA + FUNSD—are presented in \cref{tab:combine_datasets}. Each experimental configuration independently initializes from the base Qwen2-VL-7B model.

Training with RIDGE alone achieves substantial improvements across all benchmarks. While incorporating DocVQA and FUNSD brings additional gains in some cases, these improvements are relatively modest. 
The effectiveness of RIDGE is evident from its contribution: on FUNSD, RIDGE alone contributes to a 5.83\% improvement, and on CORD, it yields a 1.49\% improvement. Even with the addition of real-world datasets, the maximum gains only reach 7.65\% on FUNSD and 1.54\% on CORD. These results demonstrate that RIDGE can provide a strong foundation for VIE tasks.

\begin{table}[!t]
\centering
\resizebox{1\columnwidth}{!}{
    \begin{tabular}{l|c|cc}
    \toprule
    \multirow{2}{*}{Methods} & \textbf{SROIE\textsuperscript{--}}          & \multicolumn{2}{c}{\textbf{EPHOIE}}             \\
                             & \textit{ANLS\%}                   & \textit{Acc\%}              & \textit{ANLS\%}              \\ \hline
    Qwen2-VL-7B~\cite{Qwen2VL}              & 97.50                    & 76.91                  & 86.52                  \\
    +\textit{RIDGE}                   & 97.74   (+0.24)          & 77.89   (+0.98)        & 87.79   (+1.27)        \\
    +\textit{RIDGE}+\textit{RIDGE-DS}       & \textbf{98.05   (+0.55)} & \textbf{80.91 (+4.00)} & \textbf{89.82 (+3.30)} \\
    \bottomrule
    \end{tabular}
}
\caption{Performance improvements in closed-set VIE after applying domain-specific document generation. "\textit{RIDGE}" denotes the general open-set documents generated by RIDGE. "\textit{RIDGE-DS}" denotes domain-specific documents generated by RIDGE.}
\vspace{-10pt}
\label{tab:domain-specific}
\end{table}

\subsection{Domain-Specific Document Generation}
\label{sebsec:domain_specific}
To address the challenges in closed-set VIE discussed in \cref{subsec:finetune_MLLMs}, we generate additional synthetic 
data that directly align with specific benchmark's pre-defined key sets. In this experiment, we focus on SROIE and EPHOIE. For these benchmarks, we prompt GPT-4o~\cite{ray2023chatgpt} to generate text content that mimics their respective domains: receipts and exam cover pages. More details can be found in the supplemental. Subsequently, our CLGM tackles the layout design. Examples of generated images are shown in ~\cref{fig:example}(b)(c). 

We conduct experiments using Qwen2-VL, initializing from an enhanced version trained with RIDGE alone (as discussed in \cref{subsec:finetune_MLLMs}) for better comprehension of document structure. We perform separate training for each benchmark using approximately 700 SROIE-styled documents and 800 EPHOIE-styled documents respectively. The performance gains after training on these domain-specific documents are presented in \cref{tab:domain-specific}. The results show significant performance improvements on both benchmarks. Notably, on EPHOIE, we observe substantial improvements with accuracy increasing by 4.00\% and ANLS rising by 3.30\%. These improvements demonstrate the effectiveness of our domain-specific document generation and validate the reliability of 
data generated by RIDGE.

\subsection{Applied RIDGE on LayoutLMv3}
\label{sebsec:LayoutLmv3}
In addition to analyzing entity relations, Semantic Entity Recognition (SER), which focuses on classifying entities into semantic categories, is also a common subtask of VIE.
In this experiment, we utilize LayoutLMv3\textsubscript{\textit{BASE}} \cite{huang2022layoutlmv3}, which has been widely adopted for VIE tasks, with few-shot learning to verify the quality of the semantic category annotations in RIDGE and its effectiveness on SER task.
We discuss whether introducing RIDGE for pre-training prior to few-shot fine-tuning on real images can boost performance on real datasets.
Following LayoutLMv3, a linear classifier is added to adapt the model for SER downstream task.

We evaluate on FUNSD and XFUND-ZH, as RIDGE adheres to the same semantic category labels.
We conduct experiments with and without RIDGE pre-training, followed by few-shot fine-tuning on the training set and evaluation on the testing set of each dataset. 
As shown in \cref{tab:SER}, the use of RIDGE consistently enhances model performance across all few-shot settings.
Notably, RIDGE demonstrates effective zero-shot adaptation, suggesting the reliability of its semantic category labels 
for real-world document simulation.

\begin{table}[!t]
\centering
\resizebox{1\columnwidth}{!}{
    \begin{tabular}{c|c|c|c|c|c}
    \toprule
    \textbf{Pre-training Data} & \textbf{Fine-tuning Data} & \textbf{0-shot} & \textbf{1-shot} & \textbf{5-shot} & \textbf{10-shot} \\
    \hline
    - & \multirow{2}{*}{FUNSD} & - & 45.42 & 72.45 & 71.03 \\
    RIDGE &  & \textbf{62.77} & \textbf{71.07} & \textbf{73.25} & \textbf{74.79} \\
    \hline
    - & \multirow{2}{*}{XFUND-ZH} & - & 52.02 & 76.39 & 81.18 \\
    RIDGE &  & \textbf{69.75} & \textbf{70.91} & \textbf{78.57} & \textbf{83.02} \\
    
    \bottomrule
    \end{tabular}
}
\caption{Few-shot learning with LayoutLMv3 for SER task. \textit{x-shot} specifies the number of fine-tuning images.} 
\vspace{-5pt}
\label{tab:SER}
\end{table}

\subsection{Ablation Study}
To assess the effectiveness of Hierarchical Structure Learning, we conduct an ablation study comparing model performance between training with classical VIE tasks alone and training with both VIE and Hierarchical Structure Learning. To ensure a fair comparison, we train both settings for the same number of iterations (800 steps) to evaluate the efficiency and effectiveness of performance gains. We conduct this study on FUNSD, XFUND-ZH, and CORD, with F1 scores reported in \cref{tab:ablation}. The first row (\textit{r1}) presents the baseline performance of Qwen2-VL-7B without any additional training. While the majority of performance gains come from VIE data generated by RIDGE (as shown by the larger improvements from \textit{r1} to \textit{r2} than those from \textit{r2} to \textit{r3}), incorporating Hierarchical Structure Learning further improves performance, yielding additional gains of 1.61\% on FUNSD, 2.52\% on XFUND-ZH, and 0.70\% on CORD.

\begin{table}[]
\centering
\resizebox{1\columnwidth}{!}{
    \begin{tabular}{c|c|ccc}
    \toprule
    \textbf{} & \begin{tabular}[c]{@{}c@{}}Hierarchical Structure\\   Learning\end{tabular} & FUNSD          & XFUND-ZH      & CORD           \\ \hline
    \textit{r1}        & Qwen2-VL-7B~\cite{Qwen2VL}                                                                      & 59.89          & 62.08          & 82.71          \\ \hline
    \textit{r2}        &                                                                                  & 64.87          & 67.32          & 83.77          \\
    \textit{r3}        & \checkmark                                                                                & \textbf{66.48} & \textbf{69.84} & \textbf{84.47} \\
    \bottomrule
    \end{tabular}
}
\caption{Ablation study on Hierarchical Structure Learning. \textit{r1} represents the baseline model without additional training, \textit{r2} is trained solely on VIE, and \textit{r3} adopts both VIE and Hierarchical Structure Learning.}
\label{tab:ablation}
\end{table}

\subsection{Interpretability}
\begin{figure}[!t]
\centering
\includegraphics[width=1.\linewidth]{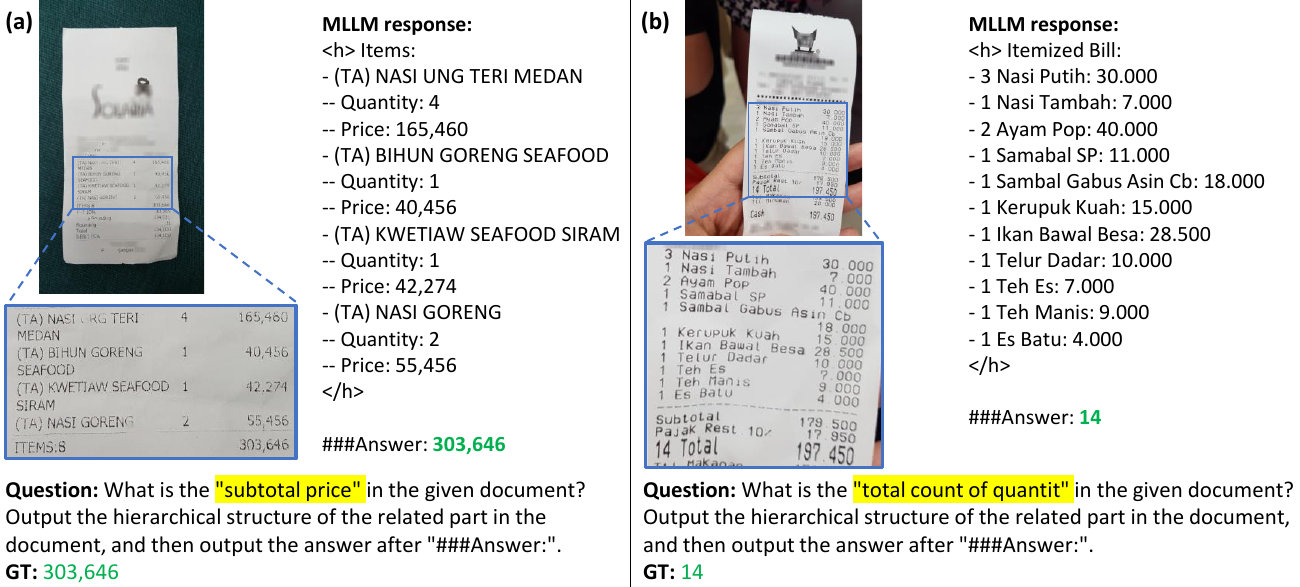}
  \caption{Interpretability brought by VIE with Hierarchical Structure Parsing.}
  \vspace{-10pt}
  \label{fig:VIE_HSP}
\end{figure}

Since we apply the VIE with HSP task during training, the fine-tuned Qwen2-VL acquires the ability to construct related information from documents in a hierarchical format before responding with the final answer. As shown in \cref{fig:VIE_HSP}(a), before extracting the "subtotal price," the model first outputs each item's quantity and price. Also, in \cref{fig:VIE_HSP}(b), before extracting the "total count of quantit," the model lists each item's information first. The process of hierarchical structure parsing not only helps the model think clearly but also provides users with detailed information related to the target query in a highly structured format.


%% file: sec/table_comb_RIDGE.tex
\begin{table*}[!ht]
\centering
\renewcommand{\arraystretch}{0.98}
\tabcolsep=0.08cm
\small
\resizebox{0.95\linewidth}{!}{
    \begin{tabular}{l|l|l|l|ll|l}
    \toprule
    \multirow{2}{*}{Dataset Combinations}   & \multicolumn{1}{c|}{\textbf{FUNSD}} & \multicolumn{1}{c|}{\textbf{CORD}}  & \multicolumn{1}{c|}{\textbf{CORD\textsuperscript{--}}}  & \multicolumn{2}{c|}{\textbf{POIE}}                                 & \multicolumn{1}{c}{\textbf{SROIE\textsuperscript{--}}}  \\
                               & \multicolumn{1}{c|}{\textit{F1 \%}} & \multicolumn{1}{c|}{\textit{F1 \%}} & \multicolumn{1}{c|}{\textit{ANLS \%}} & \multicolumn{1}{c}{\textit{Acc \%}} & \multicolumn{1}{c|}{\textit{ANLS \%}} & \multicolumn{1}{c}{\textit{ANLS \%}} \\ \hline
    Qwen2-VL-7B~\cite{Qwen2VL}                & 59.89                      & 82.71                      & 80.40                        & 93.51                      & 96.01                        & 97.50                       \\
    +RIDGE                    & 65.72   (+5.83)            & 84.20 (+1.49)              & 83.71   (+3.31)              & \textbf{94.45}   (+0.94)            & 96.71   (+0.70)              & 97.64   (+0.14)             \\
    +RIDGE+DocVQA           & 66.21   (+6.32)            & 84.06 (+1.35)              & \textbf{85.66}   (+5.26)              & 94.08   (+0.57)            & 96.67   (+0.66)              & \textbf{97.72}   (+0.22)             \\
    +RIDGE+DocVQA+FUNSD & \textbf{67.54}   (+7.65)            & \textbf{84.25} (+1.54)              & 85.57   (+5.17)              & 94.11   (+0.60)            & \textbf{96.72}   (+0.71)              & 97.71   (+0.21)            \\
    \bottomrule
    \end{tabular}
}
\caption{Comparison of fine-tuning MLLMs with RIDGE and real-world datasets. The values in parentheses show the performance gains compared to the baseline Qwen2-VL-7B model.}
\vspace{-10pt}
\label{tab:combine_datasets}
\end{table*}

%% file: sec/limitation.tex
\section{Limitation}
\label{sec:limitation}

In this work, we train RIDGE mainly on form-like images and demonstrate its adaptability. 
Nevertheless, the generalization of RIDGE to significantly distinct document types may require additional training to achieve optimal performance, which is also our future work.

%% file: sec/5_conclusion.tex
\section{Conclusion}
\label{sec:conclusion}


We propose RIDGE, a novel relation-rich visual document generator that leverages the powerful content-layout understanding 
of LLMs to generate meaningful synthetic document images. 
RIDGE is the first work that can automatically generate annotated relation-rich visual documents, which facilitates the training of visual document analysis models.
Additionally, we introduce Hierarchical Structure Learning to enhance models' 
comprehension of the inherent hierarchical structure in document layouts.
Through experiments on various VIE benchmarks, RIDGE can boost the performance of existing document understanding models and increase interpretability for practical applications.

%% file: macros.tex
\newcommand{\Real}{\mathbb{R}}
\newcommand{\textprompt}{\mathbf{t}}
\newcommand{\img}{\mathbf{I}}
\newcommand{\mask}{\mathbf{M}}
\newcommand{\Ppos}{P^{+}}
\newcommand{\Pneg}{P^{-}}
\newcommand{\Bbox}{\mathrm{BBox}}
\newcommand{\QueryBbox}{\mathrm{QueryBbox}^{\mathrm{(MLLM)}}}
\newcommand{\QueryPoints}{\mathrm{QueryPoints}^{\mathrm{(MLLM)}}}
\newcommand{\QueryIsInside}{\mathrm{QueryIsInside}^{\mathrm{(MLLM)}}}
\newcommand{\SamplePoints}{\mathrm{SamplePoints}}
\newcommand{\SAM}{\mathrm{SAM}}
\newcommand{\IsInside}{\mathrm{IsInside}}

\newcommand{\SAMPromptGT}{\mathrm{SAM\_Prompt}^{\mathrm{(GT)}}}
\newcommand{\SAMPromptMLLM}{\mathrm{SAM\_Prompt}^{\mathrm{(MLLM)}}}

\makeatletter
\renewcommand\paragraph{\@startsection{paragraph}{4}{\z@}%
    {.5em}%
    {-1em}%
    {\normalfont\normalsize\bfseries}}
\makeatother

\newcommand{\compress}{\vspace{-3.15mm}}
\newcommand{\smallcompress}{\vspace{-2.1mm}}
\newcommand{\smallsmallcompress}{\vspace{-1.05mm}}

%% file: sec/X_suppl.tex
\clearpage
\setcounter{page}{1}
\maketitlesupplementary
\appendix 

\begin{CJK*}{UTF8}{gkai}
\begin{table*}[!t]
\centering
    \begin{tabular}{c|l}
        \toprule
        \textbf{Benchmark} & \textbf{Question}                                                                                                                                                           \\ 
        \specialrule{1.2pt}{0pt}{1pt}
        FUNSD     & What is the content in the   "key" field? Directly output the answer.                                                                                              \\ \hline
        XFUND-ZH  & \begin{tabular}[c]{@{}l@{}}"key"的内容是什么？请直接回答答案。\\ (Translation: What is the content in the   "key" field? Directly output the answer.)\end{tabular}          \\ \hline
        CORD      & What is the "key"? Directly output the answer.                                                                                                                     \\ \hline
        EPHOIE    & \begin{tabular}[c]{@{}l@{}}这张考卷的"key"是什么？请直接回答答案。\\ (Translation: What is the "key" in   the given test   paper?   Directly output the answer.)\end{tabular} \\ \hline
        POIE      & What is the "key"? Directly   output the answer.                                                                                                                   \\ \hline
        CORD\textsuperscript{--}     & What is the "key" in the given document? Directly   output the answer.                                                                                             \\ \hline
        SROIE\textsuperscript{--}    & What is the "key" in the given document? Directly   output the answer.   \\                \bottomrule                                    
    \end{tabular}
\caption{Question prompts used for each benchmark.}
\label{tab:prompts}
\end{table*}
\end{CJK*}

\section{Content Generation Prompts}
\label{sec:prompts}
\begin{figure*}[!t]
\centering
\includegraphics[scale=0.65]{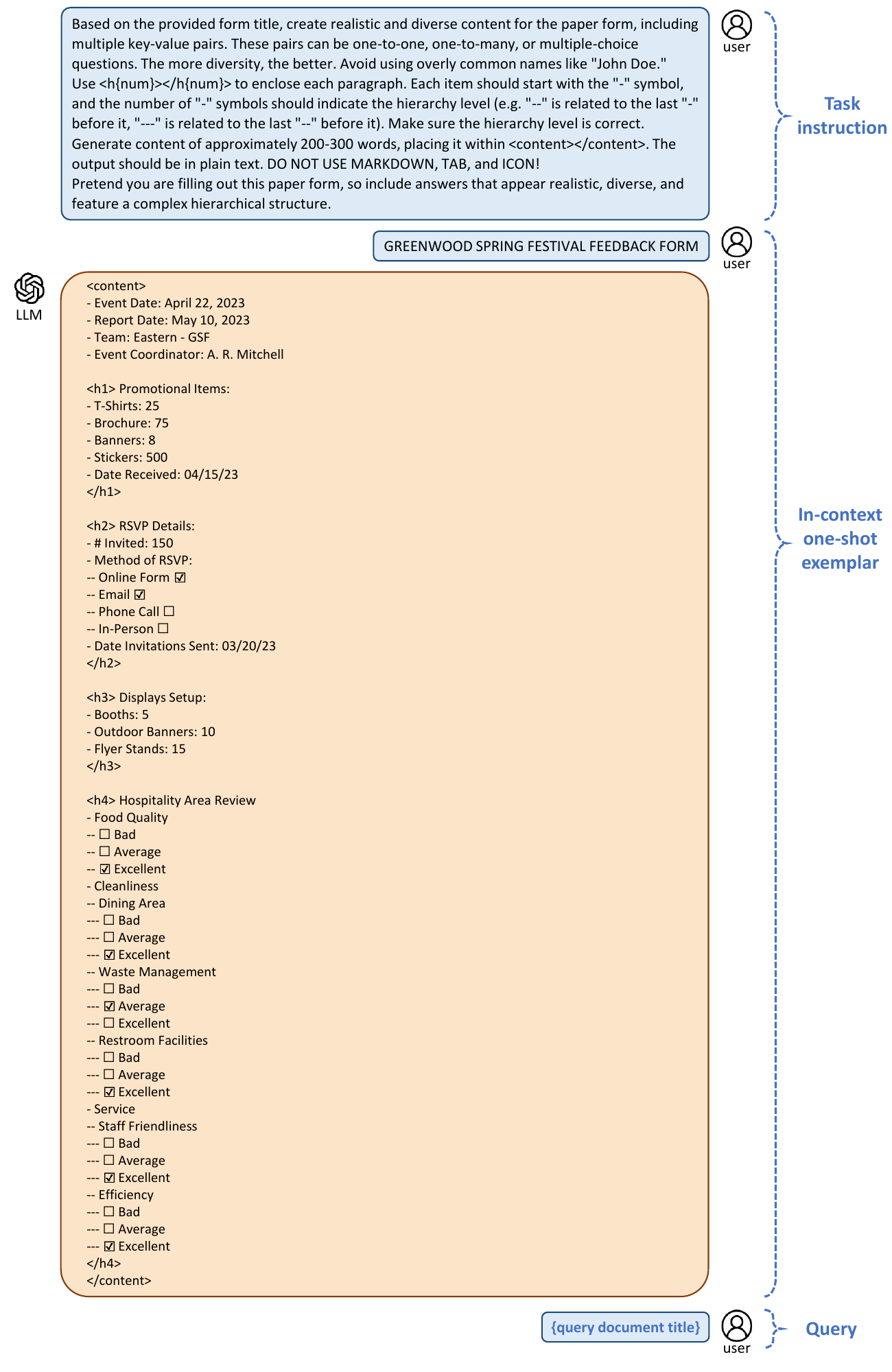}
  \caption{The prompt for guiding GPT-4o-mini in generating HST-format content for a given document title.}
  \label{fig:pmt_general}
\end{figure*}

\begin{figure*}[!t]
\centering
\includegraphics[scale=0.75]{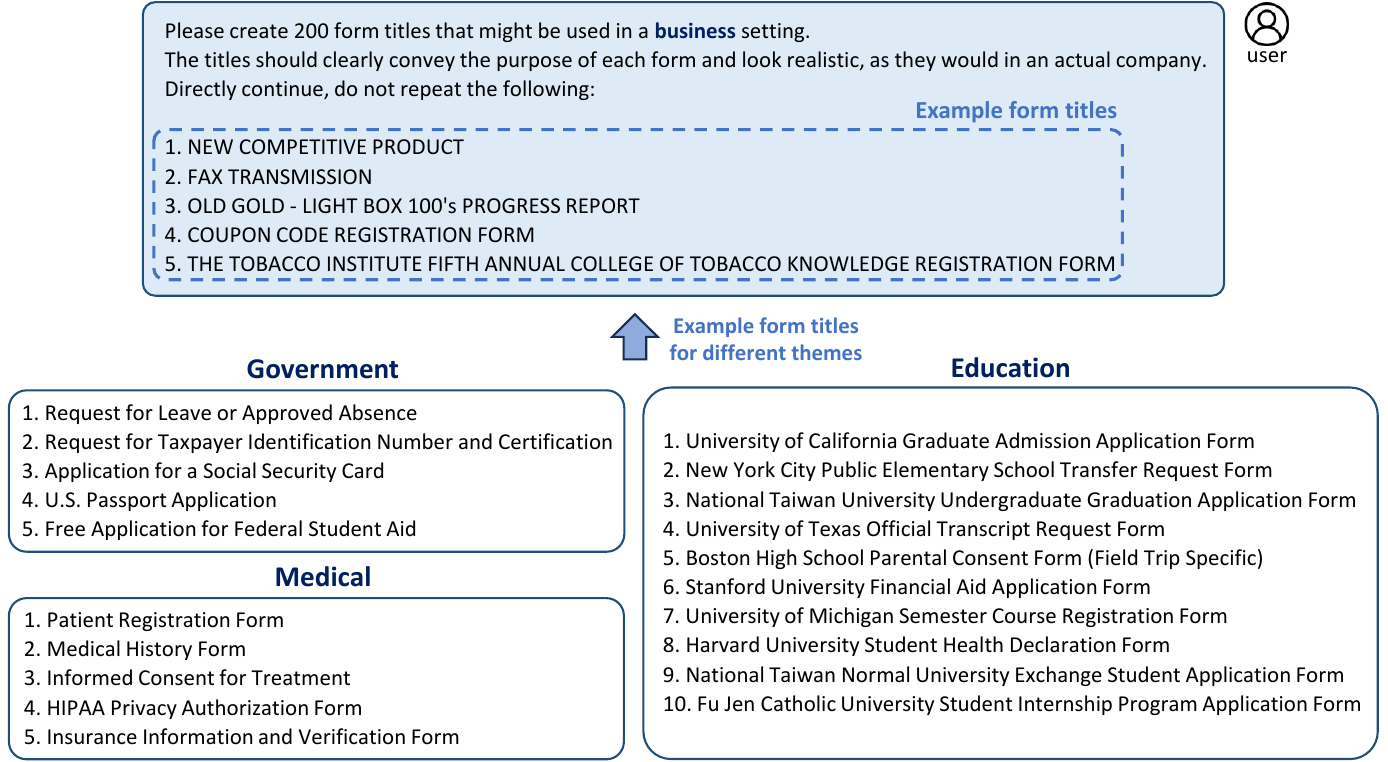}
  \caption{Prompts for generating form titles across various themes, including business, government, education, and medical.}
  \label{fig:header_creation}
\end{figure*}

For content generation in  ~\cref{subsec:content_generation}, we use the prompt shown in ~\cref{fig:pmt_general} to guide GPT-4o-mini~\cite{ray2023chatgpt} in generating HST-format content for a given document title. Additionally, we use GPT-4o-mini to generate numerous document titles covering themes such as business, government, education, and medical. The prompt is demonstrated in ~\cref{fig:header_creation}.

\begin{figure*}[!t]
\centering
\includegraphics[scale=0.65]{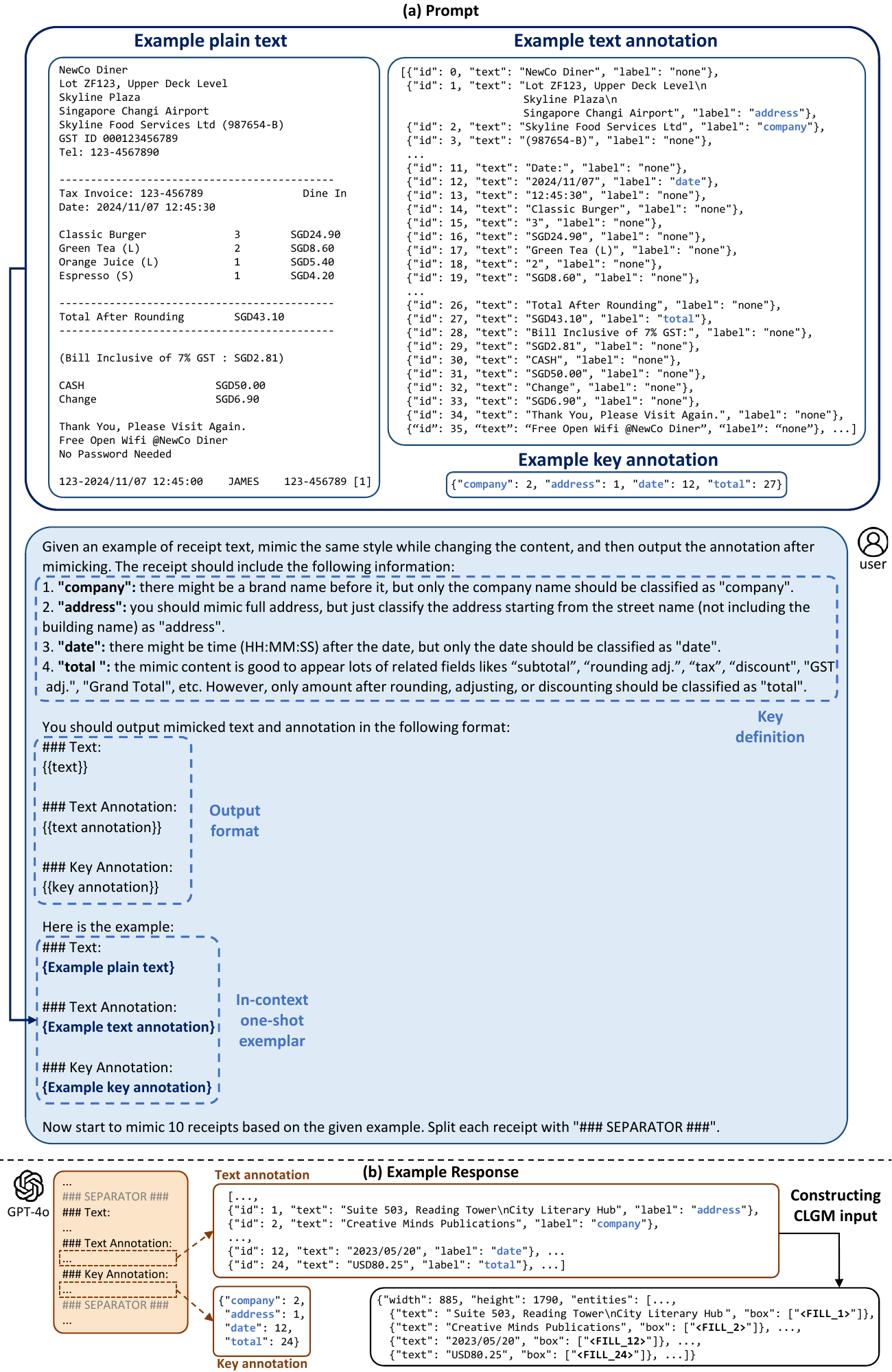}
  \caption{The prompt for generating SROIE-styled content. (a) Prompt for GPT-4o. (b) Example response from GPT-4o.}
  \label{fig:pmt_sroie}
\end{figure*}

\begin{figure*}[!t]
\centering
\includegraphics[scale=0.8]{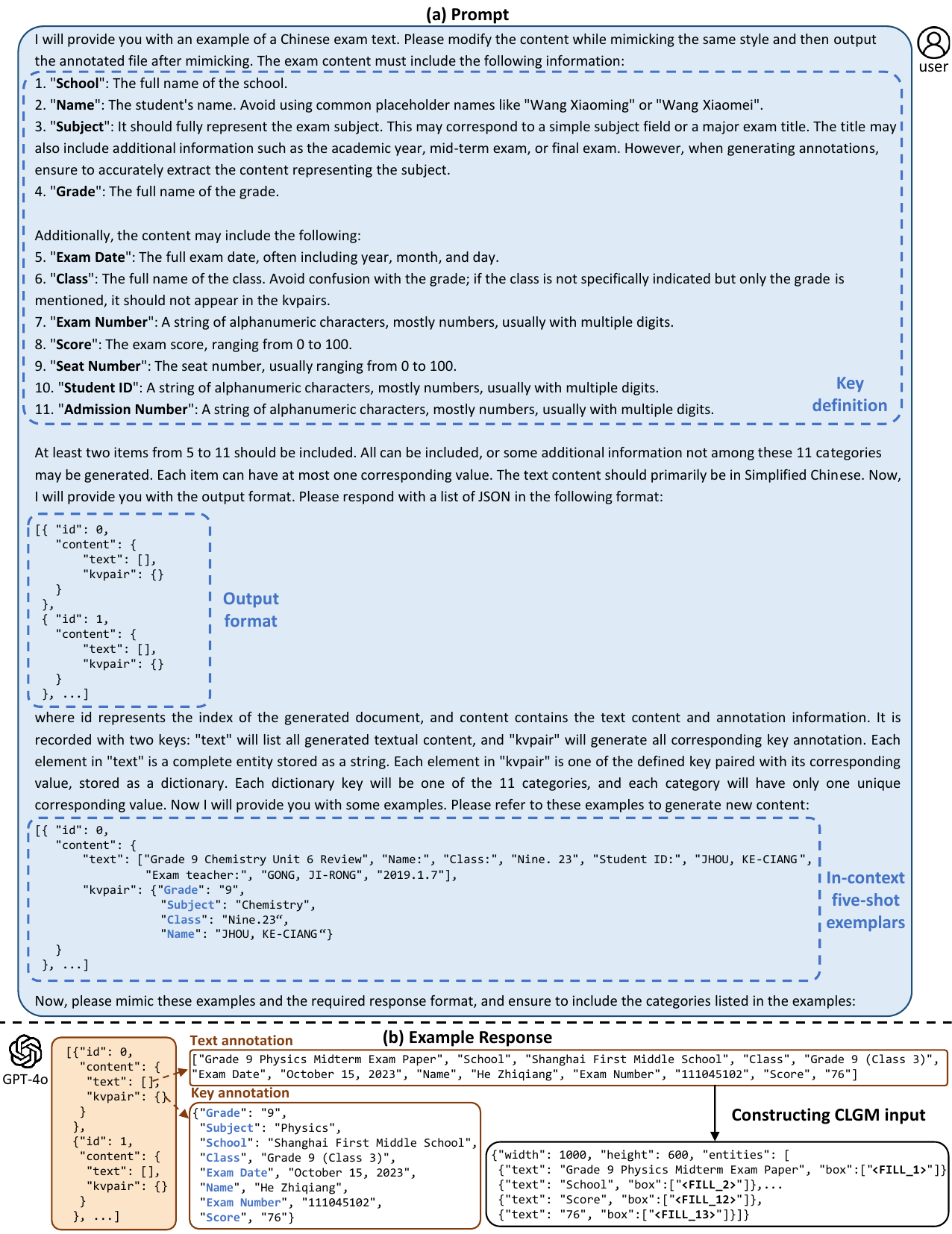}
  \caption{The prompt for generating EPHOIE-styled content. (a) Prompt for GPT-4o. (b) Example response from GPT-4o.}
  \label{fig:pmt_ephoie}
\end{figure*}

For domain-specific document generation in ~\cref{sebsec:domain_specific}, we use GPT-4o to generate document content for the domains of SROIE~\cite{huang2019icdar2019} (receipts) and EPHOIE~\cite{wang2021towards} (exam cover pages). Key-value annotations are also provided by GPT-4o using properly designed prompts. The prompt for SROIE is shown in ~\cref{fig:pmt_sroie}. The prompt for EPHOIE is originally designed in Chinese; its English translation is presented in ~\cref{fig:pmt_ephoie}. Both prompts first describe the task and the definitions of pre-defined keys, then describe the output format, and finally provide in-context demonstrations for GPT to reference. After obtaining the text content and annotations from GPT-4o, the text content is transformed into the CLGM input format, which then handles the layout design for these documents.

\section{Training Setup}
\label{sec:train_setup}
CLGM is initialized from LLaMA-3.1-8B~\cite{dubey2024llama} and is fine-tuned using LoRA~\cite{hu2022lora} adapter. The LoRA configuration is set as follows: LoRA rank and LoRA alpha are 64, LoRA dropout is 0.1. LoRA is applied to all the linear layers. The model is trained on 4 NVIDIA V100 GPUs (32GB) with float16 precision. We use paged AdamW 32-bit with weight decay of 0.001 as the optimizer. The total batch size is set to 128. We adopt a cosine annealing learning rate scheduler with a learning rate of 3e-4 and 3\% warmup ratio over 4,800 iterations. The total training takes about 13 days.

\section{Evaluation Setup}
\label{sec:eval_setup}
This section elaborates on (1) how VIE benchmarks are transformed into QA formats to evaluate the performance of MLLMs, (2) the ground truth correction for SROIE\textsuperscript{--}, and (3) the implementation details for all the document understanding models used in experiments in \cref{sec:experiments}. 

\subsection{QA for VIE benchmarks}
When leveraging LLMs/MLLMs for VIE tasks, annotations are transformed into a question-answering format, such as \{Q: What is the "key" in the given document? A: "value"\}. We provide the question prompts used for querying each benchmark in ~\cref{tab:prompts}. All experiments with MLLMs follow this setup, and their performance is reported accordingly. However, the performance of LayoutLLM~\cite{layoutllm} is taken from its original paper, as the model checkpoint is not publicly available for testing.

\subsection{Ground Truth Correction for SROIE\textsuperscript{--}}
\begin{figure*}[!b]
\centering
\includegraphics[scale=0.6]
    {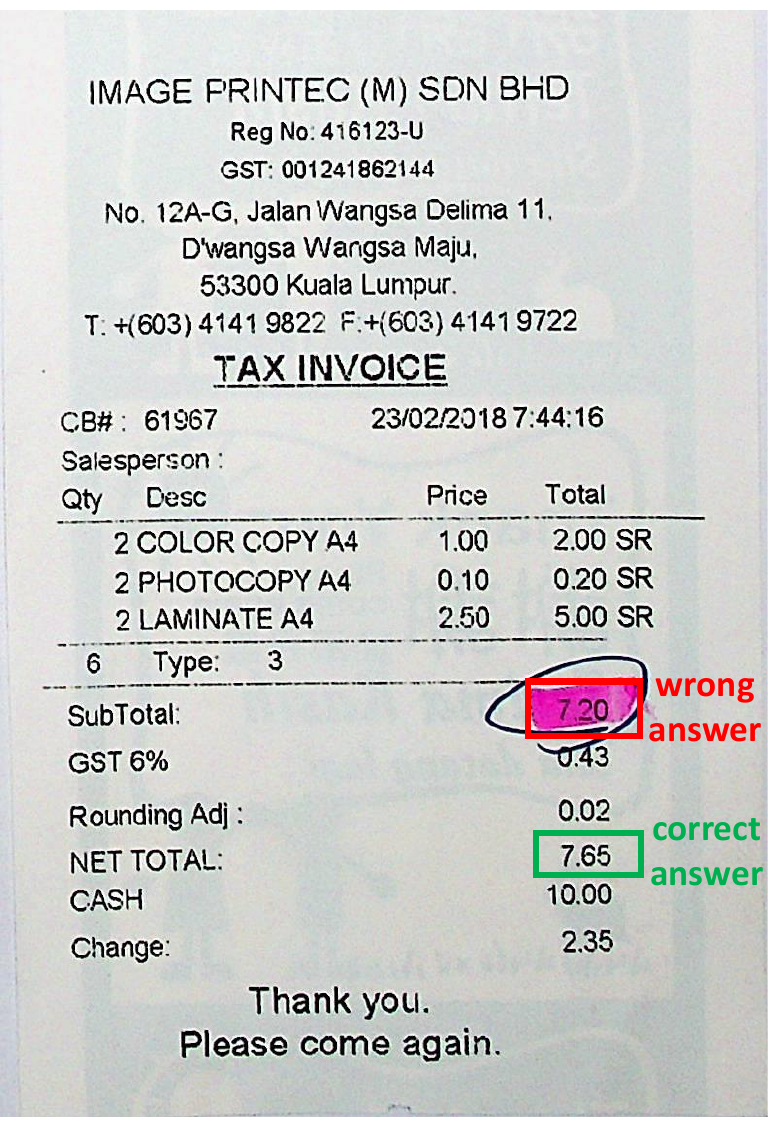}
  \caption{Wrong annotation in SROIE\textsuperscript{--}. The \textit{total} is annotated as "7.20", however, "7.65" is the correct answer.}
  \label{fig:sroie_correct}
\end{figure*}

We identified one annotation error in SROIE\textsuperscript{--} and corrected it for performance evaluation. The image is named \textit{X51005806696.jpg} in the dataset. The image with incorrect and corrected annotation is shown in ~\cref{fig:sroie_correct}.

\subsection{Implementation Details in Experiments}
This subsection elaborates on the training details of document understanding models we utilize in experiments, including MLLMs and LayoutLMv3~\cite{huang2022layoutlmv3}.

\noindent\textbf{MLLMs.} We leverage LoRA to fine-tune Qwen2-VL-7B~\cite{Qwen2VL} and LLaVA-NeXT-Mistral-7B~\cite{liu2024llavanext}. The settings for experiments in ~\cref{subsec:finetune_MLLMs} are described as follows. For LoRA configuration, we set rank to 8, alpha to 16, and dropout to 0.01, applying it to all linear layers. We use AdamW with weight decay of 0.1 as the optimizer. The total batch size is set to 128. We adopt a cosine annealing learning rate scheduler with a learning rate of 5e-6 and 10\% warmup ratio over 2 epochs. 

The settings for experiments in ~\cref{sebsec:domain_specific} are described as follows. For LoRA configuration, we set rank to 4, alpha to 8, and dropout to 0.1, applying it to all linear layers. We use AdamW with weight decay of 0.1 as the optimizer. The total batch size is set to 16. We adopt a cosine annealing learning rate scheduler with a learning rate of 1e-5 and 10\% warmup ratio over 10 epochs.

\noindent\textbf{LayoutLMv3.} The SER model used in ~\cref{sebsec:LayoutLmv3} is initialized from LayoutLMv3\textsubscript{\textit{BASE}}, with only the additional linear classifier fine-tuned. Following previous works \cite{xu2020layoutlm,xu-etal-2021-layoutlmv2,huang2022layoutlmv3}, the classifier consists of a single linear layer that maps LayoutLMv3's encoded tokens, represented by 768-dimension, to seven possible classes.
Following standard sequence labeling conventions, these classes are: \textit{begin-of-header}, \textit{begin-of-question}, \textit{begin-of-answer}, \textit{inside-of-header}, \textit{inside-of-question}, \textit{inside-of-answer}, and \textit{other}.
The dropout rate of the linear classifier is set to 0.1. The optimization process utilizes the AdamW optimizer with a learning rate of 1e-5.

\section{More Examples of Generated Images}
\label{sec:visual_results}
\begin{figure*}[!t]
\centering
\includegraphics[scale=0.82]{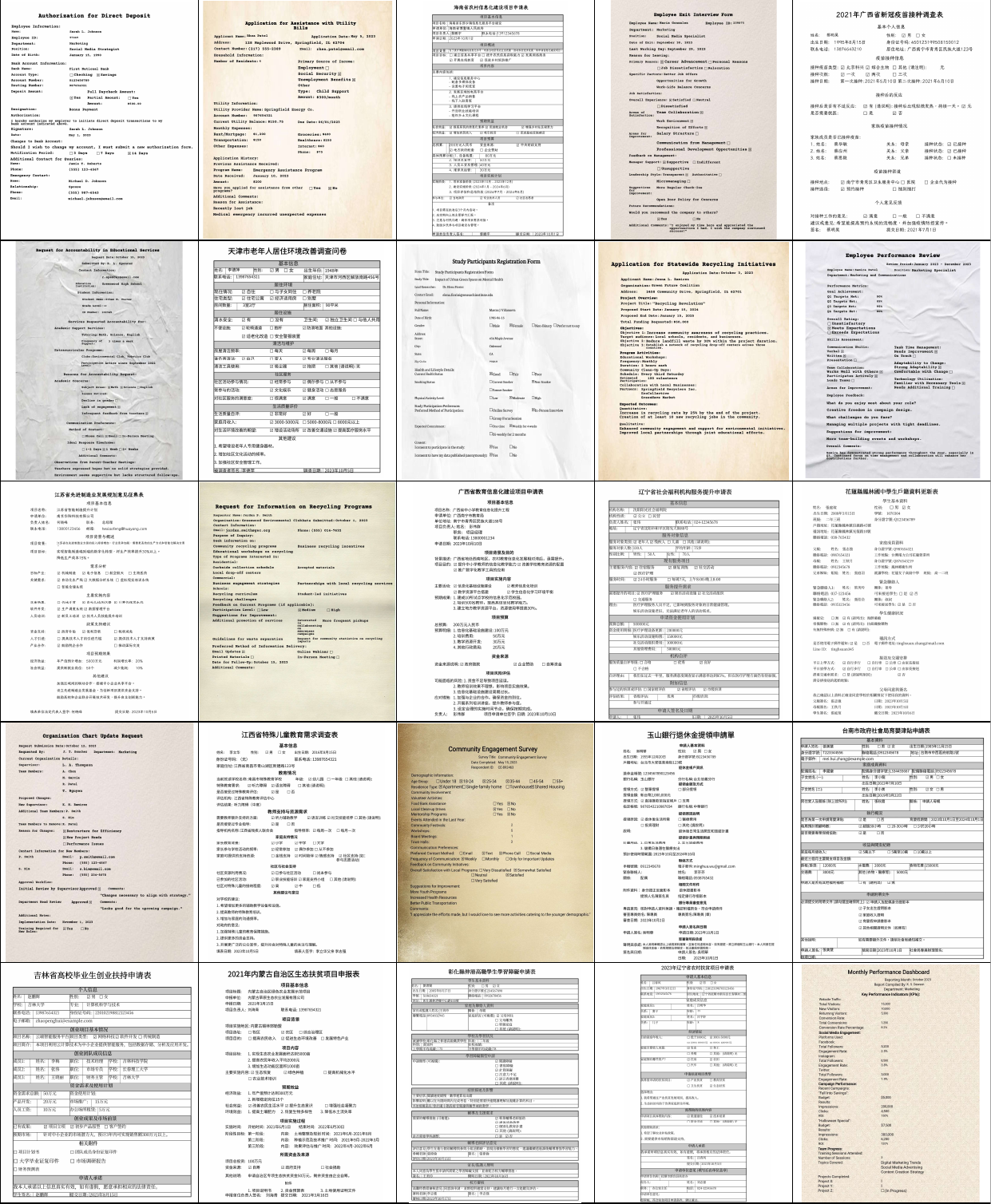}
  \caption{Examples of document images generated by RIDGE.}
  \label{fig:visualize}
\end{figure*}

We demonstrate more examples of document images generated by RIDGE in ~\cref{fig:visualize}.

\begin{table}[!t]
\centering
\tabcolsep=0.06cm
\resizebox{1\columnwidth}{!}{
\begin{tabular}{l|cccc|cccc}
\toprule
\multirow{2}{*}{}   & \multicolumn{4}{c|}{\textbf{FUNSD}}                                                                                                                                                                      & \multicolumn{4}{c}{\textbf{XFUND-ZH}}                                                                                                                                                                     \\ \cline{2-9}
          & \textit{FID{$\downarrow$}} & \textit{Over.{$\downarrow$}} & \multicolumn{1}{c|}{\textit{Align.{$\downarrow$}}} & \textit{\begin{tabular}[c]{@{}c@{}}LayoutLMv3 \\ FID{$\downarrow$}\end{tabular}} & \textit{FID{$\downarrow$}} & \textit{Over.{$\downarrow$}} & \multicolumn{1}{c|}{\textit{Align.{$\downarrow$}}} & \textit{\begin{tabular}[c]{@{}c@{}}LayoutLMv3 \\ FID{$\downarrow$}\end{tabular}} \\ \hline
Ours      & 9.59                         & 4.75                           & \multicolumn{1}{c|}{0.07}                           & 7.66                                                                               & 14.76                        & 3.00                           & \multicolumn{1}{c|}{0.01}                            & 7.45                                                                               \\ 
\textit{Real data} & 9.19                         & 2.31                           & \multicolumn{1}{c|}{0.09}                           & 5.83                                                                               & 12.81                        & 0.92                           & \multicolumn{1}{c|}{0.04}                            & 1.04     \\                          \bottomrule                                               
\end{tabular}
}
\caption{Layout and content evaluation on FUNSD and XFUND-ZH. \textit{Align.} represents Alignment. \textit{Over.} represents Overlap.}
\label{tab:layout_eval}
\end{table}

\section{Layout \& Content Evaluation.}
\label{sec:layout_eval}
In \cref{tab:layout_eval}, we follow previous works~\cite{jiang2023layoutformer++,inoue2023layoutdm,lin2023layoutprompter} using FID, Alignment, and Overlap metrics to evaluate our layout generation quality on FUNSD and XFUND-ZH datasets. 
The FID between train and test sets serves as \textit{Real data}. Comparing it to the FID between test set and synthetic layouts, the small differences show that CLGM can produce realistic layouts.
To evaluate 
both \textit{content \& layout} concurrently, 
we use LayoutLMv3 as a feature extractor due to its joint understanding of textual and spatial information.
We synthesize document images based on texts from each dataset's test set. 
LayoutLMv3 FID is computed using CLS token embeddings.
With no 
studies for joint evaluation of content and layout, our results serve as a baseline for future studies.

\section{One-Stage \& Two-Stage Generation.}
\label{sec:one_stg}
\begin{figure*}[!t]
\centering
\includegraphics[scale=0.85]{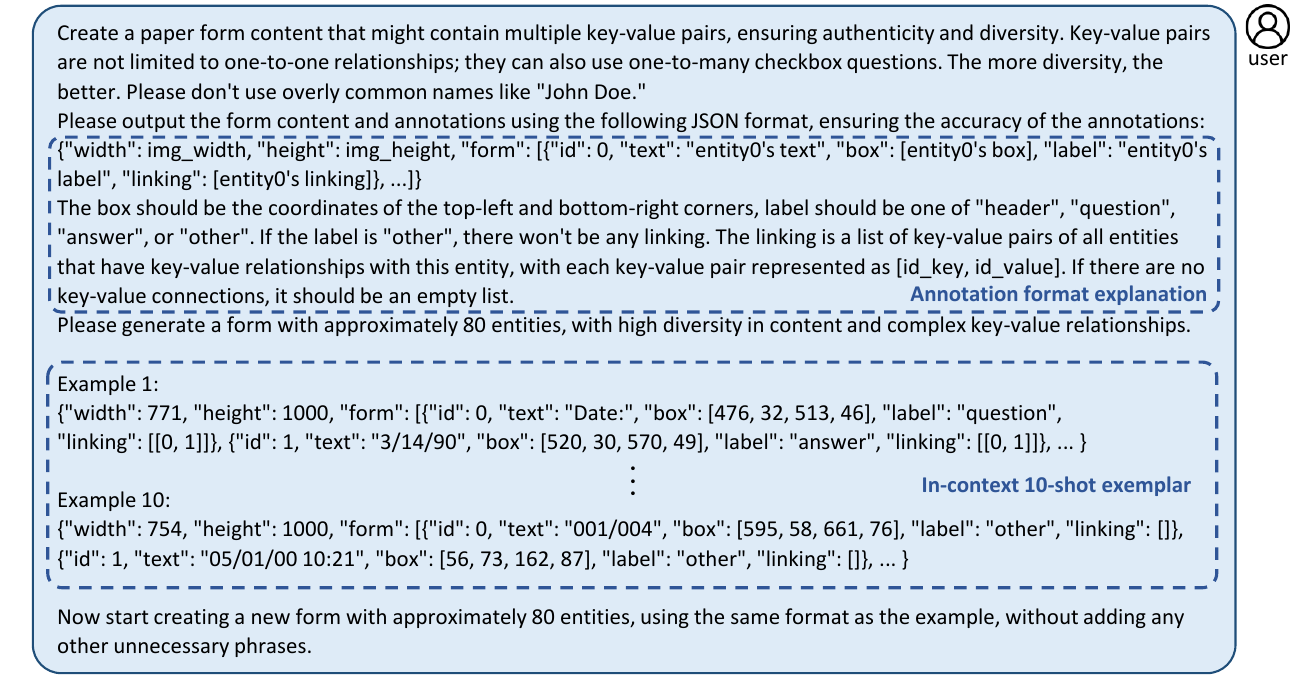}
  \caption{The prompt for one-stage generation of relation-rich visual document.}
  \label{fig:one_stg_pmt}
\end{figure*}

\begin{figure*}[!t]
\centering
\includegraphics[scale=0.15]{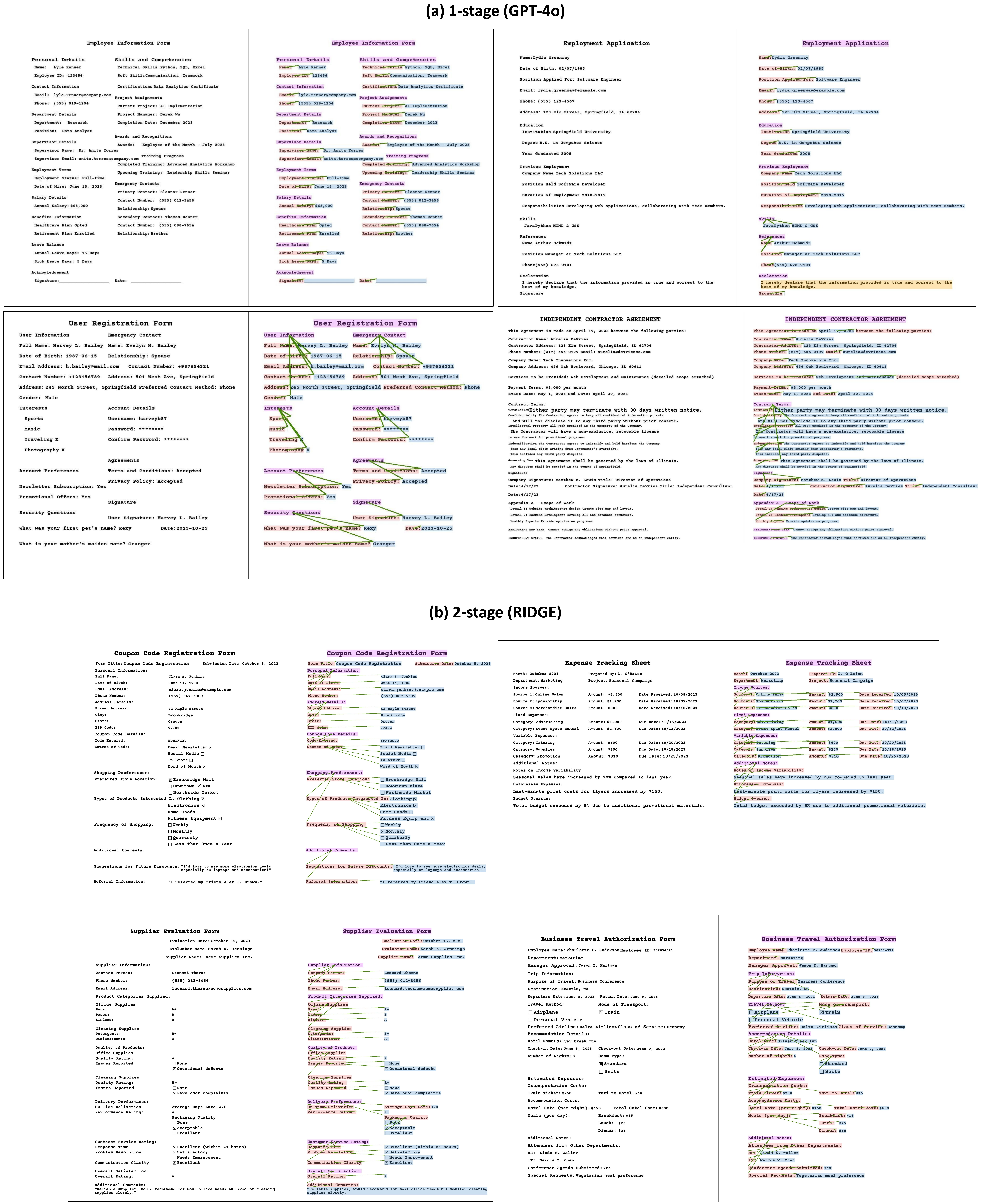}
  \caption{Qualitative comparison of generated English documents between (a) one-stage generation using GPT-4o and (b) two-stage generation using RIDGE.}
  \label{fig:one_stg_EN}
\end{figure*}

\begin{figure*}[!t]
\centering
\includegraphics[scale=0.15]{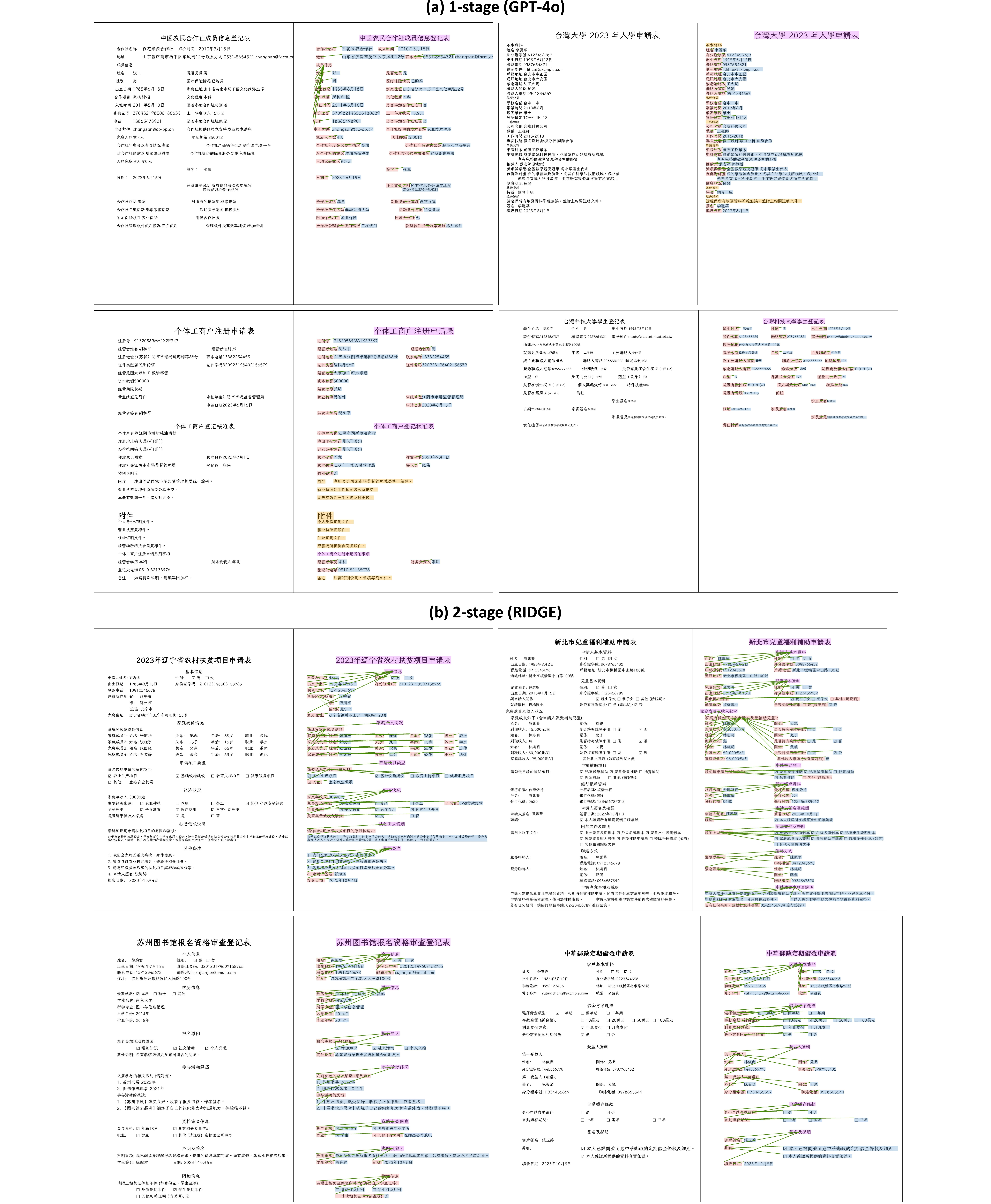}
  \caption{Qualitative comparison of generated Chinese documents between (a) one-stage generation using GPT-4o and (b) two-stage generation using RIDGE.}
  \label{fig:one_stg_ZH}
\end{figure*}

In this section, we discuss the possibility of one-stage generation and provide comparisons between one-stage and two-stage generation. When approaching one-stage generation of relation-rich visual documents, which means simultaneously generating textual content, annotations (entity category and linking), and layouts, we leverage in-context learning with LLMs, rather than applying supervised fine-tuning, since existing open-set VIE datasets are scarce. We randomly choose 10 documents each from FUNSD and XFUND-ZH and use their full annotations as in-context learning exemplars. The input entity order in each exemplar is sorted by bounding box coordinates from left-top to right-bottom to mimic the natural reading order and improve LLMs' understanding. The prompt is provided in \cref{fig:one_stg_pmt}. We conduct the one-stage generation experiment using GPT-4o, with qualitative results shown in \cref{fig:one_stg_EN}(a) for English documents and \cref{fig:one_stg_ZH}(a) for Chinese documents. For comparison, the two-stage results from RIDGE are presented in \cref{fig:one_stg_EN}(b) and \cref{fig:one_stg_ZH}(b), respectively. Each example image pair presents the raw generated document on the left and its corresponding visualized annotation on the right. All results are displayed without styling operations and use the same font for fair comparison.

In our observations, these two approaches exhibit several significant differences. From a layout perspective, the one-stage method produces layouts that are more monotonous and lack proper alignment, thereby reducing readability. In contrast, the two-stage method generates diverse and complex hierarchical structures with high alignment, resulting in a clear visual hierarchy and scannable structure. Regarding relation annotations, the one-stage approach predominantly generates simple one-to-one relationships, omitting necessary hierarchical relations. Nevertheless, in several examples, we observe its attempts to produce more complex relationships (e.g., one-to-many with multiple hierarchical levels), though these attempts often result in notable entity linking errors. We hypothesize that this may be attributed to the simultaneous processing of numerous entities containing substantial numerical content (including bounding box coordinates and entity id references within linking), which constitutes an excessively complex task with long sequences, hindering the model's ability to leverage its inherent knowledge. By contrast, the two-stage method generates complex linking with highly hierarchical structure.

Furthermore, we provide a quantitative analysis of relation complexity and failure rates for document generation in \cref{tab:one_stg}. 
For a fair comparison, we conduct 50 generation attempts across both one-stage and two-stage methods, each applied to English and Chinese documents separately.
In terms of relation complexity, entities generated by the one-stage method mainly fall within levels 0 and 1 with a nearly 1:1 distribution, indicating almost exclusively one-to-one relationships that lack complexity. In contrast, the two-stage method shows a more diverse hierarchical distribution with substantial presence across levels 0-3 and some in deeper levels, demonstrating richer relationship structures.
Additionally, the failure rate represents the percentage of samples that could not be successfully processed (e.g., incorrect JSON format) or contained notable linking errors. The one-stage method exhibits significantly higher failure rates than the two-stage approach. Moreover, such a high failure rate (48\% in English one-stage generation) would likely impede effective deployment in practical scenarios. 

In conclusion, two-stage generation effectively decomposes the generation of textual content with annotations and layout into separate subtasks, reducing the complexity of each task and maximizing the potential of LLMs.


\begin{table}[!t]
\centering
\tabcolsep=0.06cm
\resizebox{1\columnwidth}{!}{
\begin{tabular}{ll|cccccc|c||c}
\toprule
                                         &                  & \multicolumn{6}{c|}{hierarchical level} & \multirow{2}{*}{\begin{tabular}[c]{@{}c@{}}avg \#entities \\ per doc\end{tabular}} & \multirow{2}{*}{\begin{tabular}[c]{@{}c@{}}failure \\ rate\end{tabular}} \\
                                         &                  & 0     & 1     & 2     & 3   & 4   & 5   &                                                                                  &                                                                          \\ \hline
\multicolumn{1}{l|}{\multirow{2}{*}{EN}} & 1-stage (GPT-4o) & 18.1  & 15.1  & 0.5   & 0.0 & 0.0 & 0.0 & 34.0                                                                             & 48\%                                                                     \\
\multicolumn{1}{l|}{}                    & 2-stage (RIDGE)  & 9.1   & 15.2  & 22.1  & 9.1 & 1.3 & 0.0 & \textbf{56.8}                                                                    & \textbf{6\%}                                                             \\ \hline
\multicolumn{1}{l|}{\multirow{2}{*}{ZH}} & 1-stage (GPT-4o) & 25.9  & 26.3  & 0.7   & 0.0 & 0.0 & 0.0 & 52.9                                                                             & 20\%                                                                     \\
\multicolumn{1}{l|}{}                    & 2-stage (RIDGE)  & 9.4   & 23.0  & 38.3  & 9.4 & 2.3 & 0.1 & \textbf{82.5}                                                                    & \textbf{10\%}          \\ \bottomrule                                                 
\end{tabular}
}
\caption{Analysis of relation complexity and failure rates for document generation. The values in the \textit{hierarchical level} columns indicate the average number of entities at each hierarchical level per document.}
\label{tab:one_stg}
\end{table}

\section{Visual Information Extraction Methods.}
\label{sec:VIE_review}
Visual Information Extraction (VIE) is a crucial task within Visual Document Understanding (VDU), encompassing entity labeling and extracting relations between entities in visually-rich documents. The methodologies can be categorized into two main approaches: OCR-based and OCR-free. In OCR-based methods, early studies~\cite{9412669, Doc2Graph} employ Graph Neural Networks (GNNs) to model entity information and relationships. Several pre-trained language models~\cite{xu2020layoutlm,hong2022bros,tu2023layoutmask} emerge utilizing transformers to model text and layout interactions, with some additionally incorporating image modality~\cite{xu-etal-2021-layoutlmv2, huang2022layoutlmv3,peng2022ernie}. These models leverage self-supervised learning with large volumes of document data, enabling application to general documents. LayoutLMv3~\cite{huang2022layoutlmv3} is one of these models that is widely used and can incorporate different downstream modules for fine-tuning on various datasets or tasks. However, this approach is limited by the requirement for specific fine-tuning on each distinct dataset or task. With the emergence of LLMs and their superior generalization capabilities, DocLLM~\cite{wang-etal-2024-docllm} integrates layout information with LLMs, while LayoutLLM~\cite{layoutllm} incorporates both layout and image modalities into LLMs. In OCR-free approaches, early studies~\cite{kim2022donut,dessurt} utilize end-to-end fashion with transformers to address document understanding challenges. Recently, MLLMs have been leveraged to solve document-oriented problems.  Although document-centric models~\cite{docowl1.5,ye-etal-2023-ureader} are initially more prevalent, with advancements in MLLMs, general-purpose MLLMs~\cite{Qwen2VL,liu2024llavanext,chen2024far} achieve promising performance in document-oriented tasks. For instance, Qwen2-VL~\cite{Qwen2VL} approaches human-level performance on DocVQA, suggesting a trend toward utilizing general-purpose MLLMs in the VDU field.

\clearpage

\clearpage

\clearpage

\clearpage

